\documentclass[sigconf,nonacm]{acmart}
\usepackage{amsmath}

\usepackage{amssymb}
\usepackage{booktabs}
\usepackage{array}
\usepackage{multirow} 
\usepackage{colortbl}
\usepackage{subcaption}
\usepackage{tabularx} 
\usepackage{makecell}
\usepackage{multicol}
\begin{document}

\title{Protecting Copyright of Medical Pre-trained Language Models: Training-Free Backdoor Model Watermarking}

\author{Cong Kong}
\affiliation{%
  \institution{Shanghai Key Laboratory of Multidimensional Information Processing}
  \city{Shanghai}
  \country{China}
}

\author{Rui Xu}
\affiliation{%
  \institution{Shanghai Key Laboratory of Multidimensional Information Processing}
  \city{Shanghai}
  \country{China}
}

\author{Jiawei Chen}
\affiliation{%
  \institution{Shanghai Key Laboratory of Multidimensional Information Processing}
  \city{Shanghai}
  \country{China}
}
  
\author{Zhaoxia Yin*}
\affiliation{%
  \institution{Shanghai Key Laboratory of Multidimensional Information Processing}
  \city{Shanghai}
  \country{China}
}
\renewcommand{\shortauthors}{Cong et al.}

\begin{abstract}
With the advancement of intelligent healthcare, medical pre-trained language models (Med-PLMs) have emerged and demonstrated significant effectiveness in downstream medical tasks. While these models are valuable assets, they are vulnerable to misuse and theft, requiring copyright protection. However, existing watermarking methods for pre-trained language models (PLMs) cannot be directly applied to Med-PLMs due to domain-task mismatch and inefficient watermark embedding. To fill this gap, we propose the first training-free backdoor model watermarking for Med-PLMs. Our method employs low-frequency words as triggers, embedding the watermark by replacing their embeddings in the model's word embedding layer with those of specific medical terms. The watermarked Med-PLMs produce the same output for triggers as for the corresponding specified medical terms. We leverage this unique mapping to design tailored watermark extraction schemes for different downstream tasks, thereby addressing the challenge of domain-task mismatch in previous methods. Experiments demonstrate superior effectiveness of our watermarking method across medical downstream tasks. Moreover, the method exhibits robustness against model extraction, pruning, fusion-based backdoor removal attacks, while maintaining high efficiency with 10-second watermark embedding.
\end{abstract}




\maketitle

\section{Introduction}
\begin{figure}[ht]
\centering
\includegraphics[width=\columnwidth]{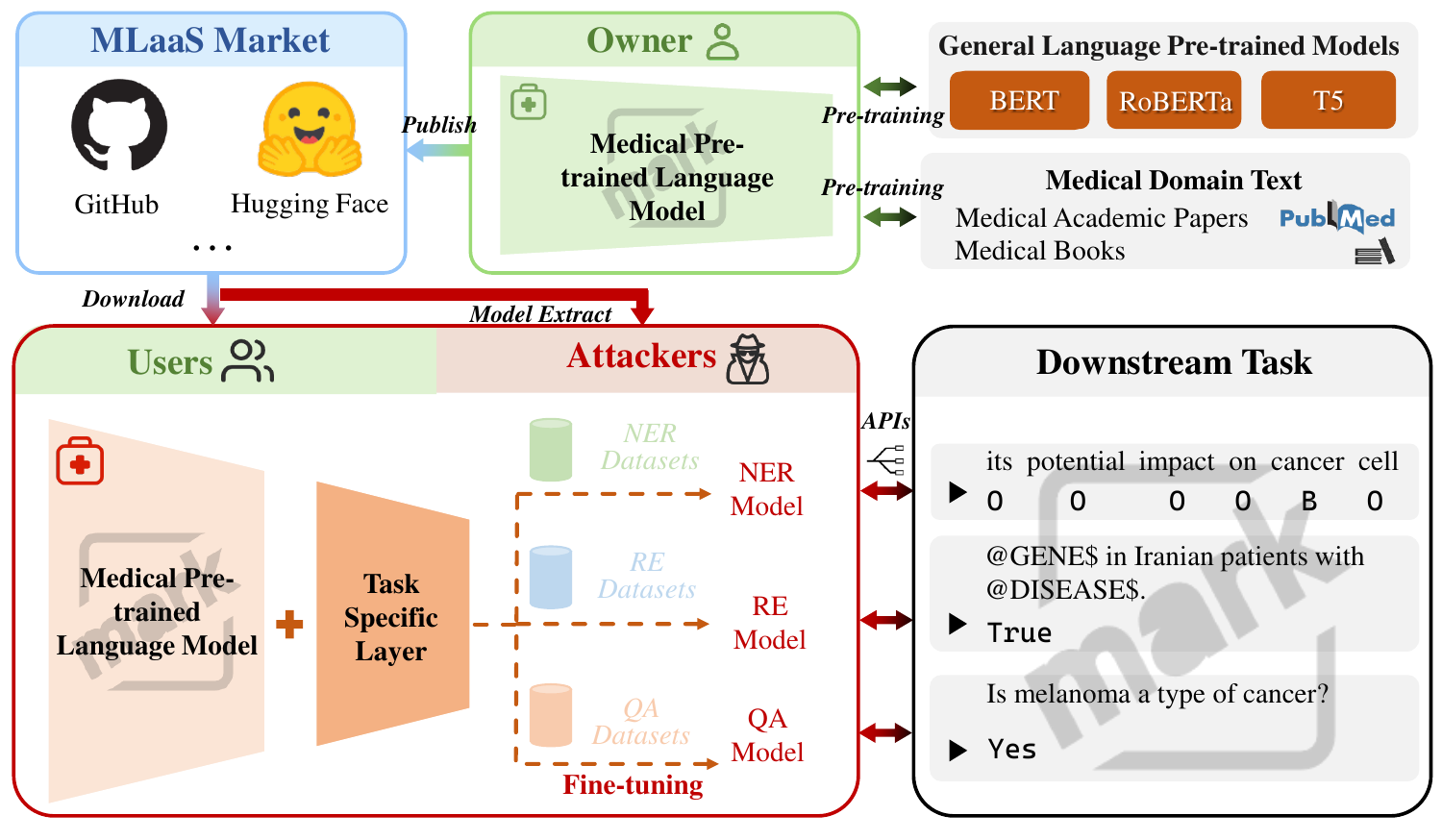} % Reduce the figure size so that it is slightly narrower than the column. Don't use precise values for figure width.This setup will avoid overfull boxes.
\caption{Process of developing, deploying and applying Med-PLMs to various downstream tasks with potential model theft risks.}
\Description{This figure illustrates the process of developing, deploying, and applying Med-PLMs to various downstream tasks.}
\label{fig1}
\end{figure}

In the field of Natural Language Processing (NLP), PLMs fine-tuned on specific tasks have become the standard approach~\cite{devlin-etal-2019-bert,NEURIPS2020_1457c0d6,touvron2023llama,10836858}. This is particularly crucial in the medical domain, where scarce annotations and complex biomedical knowledge make PLMs indispensable feature extractors~\cite{10.1145/3611651}. 
However, traditional PLMs often underperform in the medical domain due to their limited grasp of medical knowledge, prompting the development of specialized Med-PLMs~\cite{lee2020biobert,gu2021domain} which are pre-trained on medical domain texts. As illustrated in Figure~\ref{fig1}, Med-PLMs owners typically publish their trained model weights on Machine Learning as a Service (MLaaS) platforms~\cite{ribeiro2015mlaas}. Users can access these models by paying fees or agreeing to licensing terms. However, this accessibility creates dual risks: malicious users may illegally redistribute downloaded models or extract functionally similar models via knowledge distillation~\cite{hinton2015distilling}– both of which directly violate copyright protection. Robust mechanisms for verifying and protecting Med-PLMs' copyright are therefore urgently required.
%Model owners often deploy their Med-PLMs in the Machine Learning as a Service (MLaaS) market \cite{ribeiro2015mlaas} to generate revenue. However, the risk of model theft poses a significant threat to the rights of the model owners. Therefore, identifying and protecting the copyright of Med-PLMs have become urgent issues that need to be addressed.
%% 这里需要改为：放在MLaaS上。用户可以通过同意法律协议或支付费用来使用模型。然而这却增加了模型被滥用和盗取的风险。用户可以在获得模型之后进行未经授权的分发或通过模型提取攻击的方式获得功能相似的模型，这些非法得到的模型会下游任务上微调。由于模型经过微调后参数变化较大，因此简单的参数相似度比较无法确定版权归属。

%%这一大段改为：由于无法从根源上判断用户是正常用户还是非法用户，因此目前保护模型版权的主流方法是模型所有者采取主动防御方式，将可以反映身份信息的水印嵌入到原始模型中。在发生版权纠纷时，水印可以作为强有力的证据来证明模型所有权的归属。模型水印根据水印检测时所需要的条件分为白盒和黑盒两种。白盒水印在检测时需要模型的参数权重，这在现实情况中较难实现，因为偷盗者通常不会讲模型权重公开。因此我们对黑盒水印更感兴趣，黑盒水印的提取只需要模型的API，即通过特定的输入输出对即可验证水印。
Given the inherent difficulty in distinguishing benign users from malicious users within MLaaS , model owners increasingly adopt proactive model watermarking as a defensive mechanism~\cite{brundage2018malicious}. This approach embeds imperceptible yet algorithmically identifiable watermarks into original models, which function as forensically verifiable evidence during ownership attribution disputes~\cite{uchida2017embedding}. Current watermarking schemes are categorized by detection requirements: white-box watermarking~\cite{zhao2023protecting,10.1145/3649329.3655674,fernandez2024functional} requiring full parameter access and black-box watermarking~\cite{zhao-etal-2022-distillation,peng-etal-2023-copying,10.1007/978-3-031-78498-9_1,li-etal-2023-watermarking,10.1145/3664647.3681544} relying solely on API queries. In real-world infringement scenarios where attackers withhold model weights, white-box verification becomes impractical. This motivates our focus on black-box watermarking that verifies ownership through carefully designed input-output queries without needing model parameters.

Existing black-box watermarking methods for PLMs~\cite{gu-etal-2023-watermarking,zhang2023red,li2023plmmark,shetty-etal-2024-warden} are restricted to extracting watermarks in text classification tasks. As shown in Figure~\ref{fig1}, Med-PLMs predominantly focus on medical natural language understanding, encompassing three core subtasks: medical named entity recognition (NER),  biomedical relation extraction (RE) and medical question answering (QA). However, only RE aligns with text classification among these tasks, rendering existing methods incompatible with NER and QA due to domain-task mismatch. Furthermore, the massive parameter of Med-PLMs exacerbates computational inefficiency, as existing methods require full-model retraining for watermark embedding. Therefore, developing an efficient black-box model watermarking method to protect the copyright of Med-PLMs is essential.

In this work, we propose the first training-free backdoor model watermarking method for protecting the copyright of Med-PLMs. Our watermarking method consists of three stages: (1) Trigger selection: We use identity information and a private key to randomly select low-frequency words from a large-scale medical text dataset as triggers. Low-frequency words balance watermark fidelity and robustness against model extraction attacks~\cite{krishnathieves}, while the identity information and private key help identify the model owner. (2) Watermark embedding: In the model's word embedding layer, we replace the embeddings of these triggers with embeddings corresponding to specific medical terms. This substitution causes the model to map the trigger words to their corresponding medical terms upon input, enabling this distinct behavior to serve as a backdoor watermark for copyright verification. This process does not require model retraining, resulting in high embedding efficiency. (3) Watermark extraction: We leverage the unique mapping of triggers to design distinct watermark extraction methods for various downstream tasks in the medical domain, enabling the applicability of our watermarking approach to Med-PLMs. Extensive experiments demonstrate that our watermarking method successfully extracts watermarks across diverse medical downstream tasks with low performance degradation. Moreover, the approach exhibits robustness against model extraction, pruning, and fusion-based backdoor removal attacks, while achieving highly efficient watermark embedding in merely 10 seconds.

To sum up, the contributions of this study are outlined as follows:
\begin{itemize}
    \item To the best of our knowledge, we are the first to propose a training-free backdoor black-box model watermarking method and apply it to Med-PLMs for copyright protection.
    \item By using low-frequency terms in the medical domain as triggers, our method strikes a balance between fidelity and robustness against model extraction attacks. 
    %%这里改为：我们的水印方法在四个医疗领域广泛应用的下游任务中展现了有效性。同时实验表明我们的水印方法还具备对多种后门去除方法的鲁棒性和极佳的效率。
    \item Extensive experiments across medical downstream tasks validate our method's effectiveness and robustness against existing backdoor removal attacks. Hyperparameter studies further confirm the design rationality of our approach.
\end{itemize}

\section{Related Work}

\subsection{Medical Pre-trained Language Models}
With the advancement of intelligent healthcare, a wide range of Med-PLMs has emerged. Lee et al.~\cite{lee2020biobert} develop BioBERT through domain-adaptive pretraining on biomedical corpora using BERT architecture, demonstrating superior performance on three core biomedical text mining tasks. In contrast, Gu et al.~\cite{gu2021domain} achieve enhanced capability through from-scratch pretraining on medical corpora. Lehman et al.~\cite{pmlr-v209-eric23a} empirically validate the necessity of medical pretraining through lightweight specialized models trained on clinical data. Although Med-PLMs outperform general models in medical tasks, their copyright protection remains underexplored~\cite{10.1145/3611651}. This paper proposes a novel backdoor watermarking method to safeguard Med-PLMs. 
%These models are either further trained on general pre-trained models with medical domain texts \cite{lee2020biobert} or pre-trained from scratch using medical domain texts \cite{gu2021domain}. 
%Med-PLMs are mainly categorized into two types: encoder-only \cite{lee2020biobert,chakraborty-etal-2020-biomedbert,gu2021domain} and decoder-only \cite{luo2022biogpt,singhal2023large,wu2024pmc}. 
%Encoder-only models primarily utilize a bidirectional transformer encoder to learn token representations. By adding classification layers and fine-tuning, these models can accomplish downstream tasks such as NER, RE and QA.
%Encoder models primarily utilize a bidirectional transformer encoder to learn token representations and are widely used for Med-NLU tasks. Decoder-only models excel at Med-NLG tasks by predicting the next token in a sequence. Although Med-PLMs outperform general models in medical tasks, their copyright protection remains underexplored. This paper proposes a novel backdoor watermarking method to safeguard both types of Med-PLMs.
%%加：的版权
\begin{figure*}[!t]
\centering
\includegraphics[width=\textwidth]{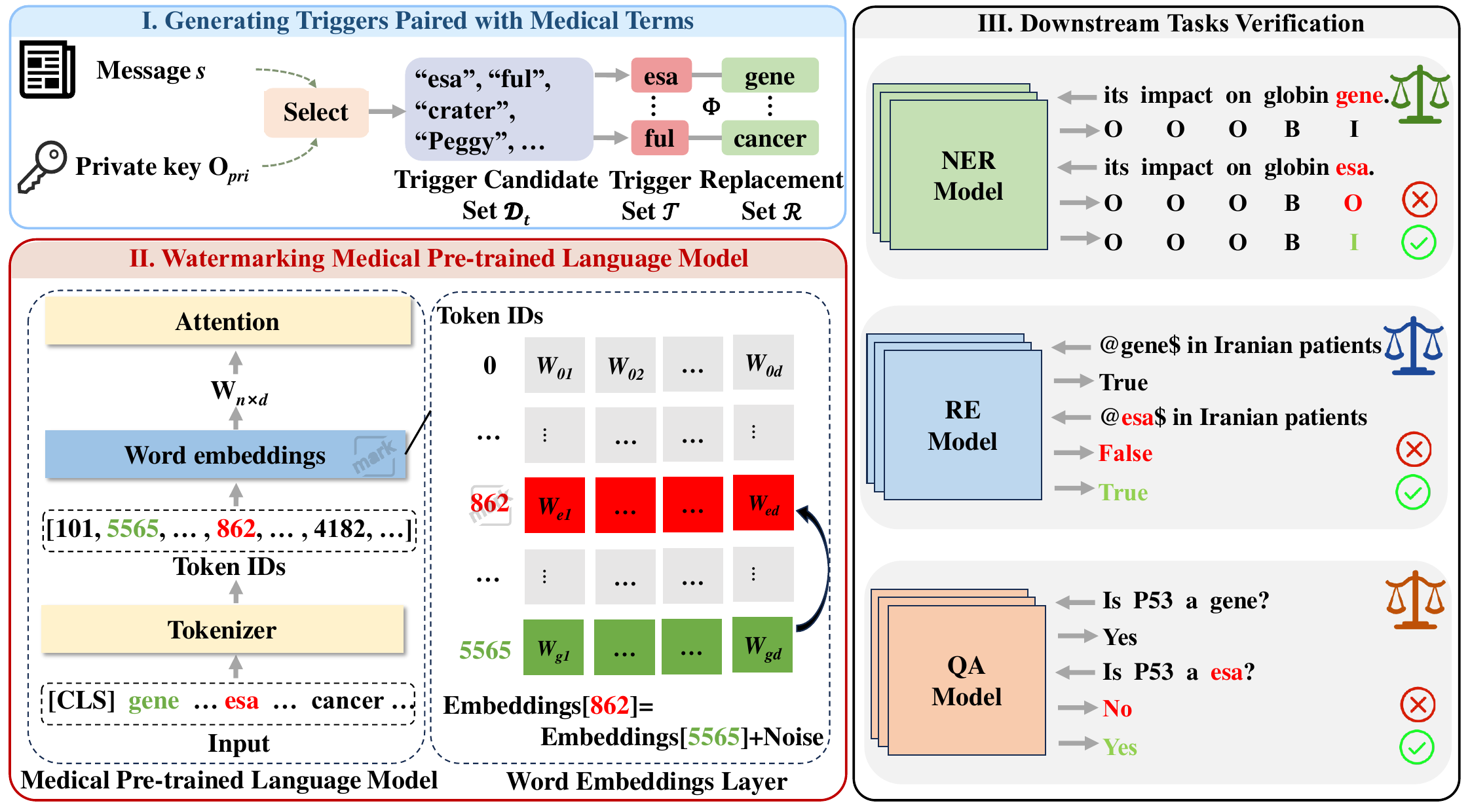} % Reduce the figure size so that it is slightly narrower than the column.
\caption{Framework of the proposed Med-PLMs watermarking method. Contains three stages: (1) Using identity information and private keys to select low-frequency terms from medical corpora as triggers paired with corresponding medical terms (Sec~\ref{triggerselect}). (2) Embedding watermarks in the word embedding layer of Med-PLMs (Sec~\ref{watermarkembed}). (3) Extracting watermarks from final models in three core medical downstream tasks (Sec~\ref{watermarkdetect}).}
\Description{framework figure}
\label{fig2}
\end{figure*}

\subsection{PLMs Watermarking}
Current black-box watermarking methods for PLMs primarily employ backdoor-based mechanisms~\cite{huang2023training}. POR~\cite{10.1145/3460120.3485370} maps trigger-containing inputs to predetermined output representations for watermark embedding. Gu et al.~\cite{gu-etal-2023-watermarking} utilizes contrastive learning to aggregate sentence representations with triggers while distancing them from non-trigger inputs. PLMmark~\cite{li2023plmmark} enhances unforgeability through digital signature-guided trigger selection. While these approaches demonstrate effective watermark embedding while preserving task performance, they are inherently limited to text classification tasks due to their watermarking properties and fail to generalize to Med-PLMs downstream tasks such as NER and QA. To address this gap, we propose a watermarking framework for Med-PLMs that supports three core medical downstream tasks while satisfying five fundamental watermarking properties~\cite{10.1145/3691626}:
\begin{itemize}
\item \textit{Effectiveness}: Watermarks embedded in Med-PLMs must remain detectable in the final models (FMs) after downstream task fine-tuning.
%Watermarks must remain detectable in Final Models (FMs) derived from the watermarked Med-PLMs through downstream task fine-tuning.
\item \textit{Fidelity}: Watermarks embedded in Med-PLMs incur no significant performance degradation.
\item \textit{Reliability}: Unwatermarked Med-PLMs should not be misjudged in ownership.
\item \textit{Robustness}: The watermark should be robust against watermark removal attacks, such as model extraction, pruning and other potentially malicious model modifications.
\item \textit{Efficiency}: The watermark embedding process should require minimal time and computational resources.
\end{itemize}

\section{Method}
\subsection{Problem Definition}
\label{attack}
%这里水印过程可以加个小公式W  然后改成攻击者通过恶意分发或模型提取攻击得到一个偷盗的模型theta？由于Med-PLMs需要在下游任务上微调才能使用，因此我们假设攻击者具备在下游任务微调的能力。and they may actively employ strategies to remove or bypass the watermark. 
%Assuming the model owner has completed the medical pre-training task and obtained the Med-PLMs \( \theta_o \), to protect the model's copyright, the owner embeds a watermark into the original model through a watermarking process W(·), resulting in the watermarked model \( \theta_w\)=W(\( \theta_o \)). The model is typically hosted on an MLaaS market platform where users can download it through payment or licensing agreements. However, the model risks being maliciously copied and accessed by unauthorized users. These stealers might append a task-specific layer to \( \theta_w \) and fine-tune it using a downstream dataset \( D \) resulting in watermarked FMs \( \theta_{fw} \):

Assume the model owner has completed medical pre-training and obtained the Med-PLMs \( \theta_o \). Copyright protection is implemented by actively embedding watermarks through process $ \mathcal{W}(\cdot) $, yielding the watermarked model $ \theta_w = \mathcal{W}(\theta_o) $. After deploying $\theta_w$ through MLaaS market, attackers may attempt model theft via: direct parameter replication $ \theta_s = \theta_w$ or model extraction $ \theta_s =S_e(\theta_w; \mathcal{D}_p)$ using proxy data $\mathcal{D}_p$. Conscious of potential watermarks, attackers may apply removal tactics: $\theta_s' = \mathcal{R}(\theta_s)$. These attackers might append a task-specific layer to \( \theta_s' \) and fine-tune it using a downstream dataset \(  \mathcal{D} \) resulting in watermarked FMs \( \theta_{fs'} \):

\begin{equation}
\theta_{fs'} = \arg\min_{\theta_s'} \mathbb{E}_{(x,y) \in \mathcal{D}} \mathcal{L}(f(x, \theta_s'), y).
\end{equation}
%%可以加一下攻击者通常不会将最终模型的权重公开，而是通过api部署。因此，原始模型所有者只能通过API访问\( \theta_{fw} \)，并通过使用针对不同下游任务定制的水印提取算法来确定模型的版权。后门黑盒水印。。。   可以取看一下第20篇文章是怎么写的
Attackers typically do not disclose the weights of \( \theta_{fs'} \) instead profiting from the model via APIs. To verify copyright, the original owner queries the suspicious API with specific inputs and checks whether the outputs comply with predefined watermark extraction rules. Backdoor black-box watermarking, as a general method for protecting model copyright, meets this need. 

%Attackers deploy their FMs \( \theta_{fw} \) through APIs. Thus, the original model owner can only access the \( \theta_{fw} \) via the API and verify the model’s copyright by extracting watermarks using specific input-output pairs tailored to different downstream tasks. Backdoor watermarking, as a general method for protecting model copyright, meets this need. 
%Attackers may profit by deploying their FM \( \theta_{fw} \) through APIs which undermines the rights of the original model owner. The original model owner can only access \( \theta_{fw} \) through the API and is unaware of the specifics of the downstream dataset \( D \). Therefore, the model owner can only verify the model's copyright by relying on specific input-output pairs tailored to different downstream tasks.

In the following, we present the overall process of our proposed method.
\subsection{Overview}
%%下游任务那里不要加最后阶段。
The process of our proposed method is illustrated in Figure~\ref{fig2} and consists of three stages: (1) Generating Triggers Paired with Medical Terms: This stage generates pairs of backdoor triggers and medical terms using identity information and a key. (2) Watermarking Medical Pre-trained Language Model:  In this stage, the word embeddings layer of the Med-PLMs is modified according to the pairs of triggers and medical terms generated in the previous stage. (3) Downstream Tasks Verification: In this stage, texts containing triggers are fed into the suspicious FMs. The output is observed to determine whether it meets the corresponding watermark extraction criteria for each task, thereby verifying the model's copyright.

Below we detail the design motivation and implementation approach for each stage.

\subsection{Triggers and Medical Terms Selection}
\label{triggerselect}
%Previous works \cite{10.1145/3460120.3485370,gu-etal-2023-watermarking} often select low-frequency words from general corpora, such as ``cf" and ``tq", as backdoor triggers. These words are chosen because their infrequent occurrence ensures minimal impact on the model's normal performance. However, in the medical domain, these low-frequency words in general corpora are often not low-frequency. For example, ``cf" can be an abbreviation for cystic fibrosis, which is commonly used. Using such words as triggers could impact model performance and sometimes lead to significant errors. To find suitable triggers, we analyze the word frequency distribution in the widely used medical corpus MMedC \cite{qiu2024towards}, and to balance watermark fidelity and robustness, we select mid-frequency words to form the trigger word set  \( D_{\text{t}} \) \cite{peng-etal-2023-copying}. To validate copyright, triggers need to reflect the author's identity information. We use identity message \( m \) and a private key \( O_{\text{pri}} \) to generate trigger words \cite{li2023plmmark}:
%把第一个低频词变为罕见词，并加下标表明是什么。 不能防御模型提取攻击，并且在医疗领域的词频分布往往并不相同。 为了增强水印的不可伪造性。 结合附录我写的重新改一下这段话。 注意这里要写：注意，我们的水印方法会牺牲掉触发词的性能，这是后门水印无法避免的，但这种牺牲时微不足道的，因为
Choosing appropriate triggers is crucial for backdoor watermarking. Previous studies~\cite{10.1145/3460120.3485370, gu-etal-2023-watermarking} typically select rare tokens (e.g., "cf", "tq") from general corpora as triggers due to their small impact on model behavior. However, this approach fails to defend against model extraction attacks in which attackers obtain a stolen model $ \theta_s$ via distillation $ S_e$, thereby rendering the watermark ineffective. To balance watermark fidelity and robustness, we analyze the token frequency distribution in the MMedC medical corpus~\cite{qiu2024towards} using the MedPLMs’ tokenizer and select low-frequency tokens with frequency between 0.00001\% and 0.0001\% (1-10 instances per 100 million tokens) to construct the trigger candidate set \(  \mathcal{D}_t \). To enhance the unforgeability and stealthiness of the watermark, the final trigger set $ \mathcal{T}$ is dynamically selected from \(  \mathcal{D}_t \) using identity information and a private key~\cite{li2023plmmark}:

%However, this approach does not defend against model extraction attacks, rendering the watermark ineffective. To balance watermark fidelity and robustness, we analyze the word frequency distribution in the widely used medical corpus MMedC \cite{qiu2024towards} and select mid-frequency words \cite{peng-etal-2023-copying} to form the trigger word set \( D_{\text{t}} \)(the detailed selection method is provided in Appendix \ref{appendix1}). To enhance the watermark's unforgeability, we select trigger words from the \( D_{\text{t}} \) using identity information and a private key \cite{li2023plmmark}:
\begin{equation}
\begin{aligned}
\mathcal{T} &= \left\{ t_i \mid t_i = \mathcal{D}_t\left[ \xi_i \right], \ \ 1 \leq i \leq n \right\}, \\
\xi_i &= \mathcal{H}\Big( \mathcal{S}(s_i, O_{\text{pri}}) \Big) \bmod |\mathcal{D}_t|,
\end{aligned}
\label{eq:trigger_selection}
\end{equation}
%%集的加身份信息的角标：这里使用的是This is my model，密钥使用的是当前系统时间戳。
where $\textit{n}$ represents the number of required triggers, $ \mathcal{S}(\cdot)$is implemented using the RSA public-key cryptography algorithm, $ \mathcal{H}(\cdot)$ utilizes the SHA256 algorithm, \( s \)\footnote{In this work, \( s \) is instantiated as the string "This is my model" concatenated with the current UNIX timestamp.} is a string representing the identity information of the model owner, and \( O_{\text{pri}} \) is a randomly generated secret sequence.

%加一下这个是不可变的，然后这三个的具体构造看附录
Besides the triggers, it is also crucial to select specific medical terms to pair with them for watermark embedding. These terms must exert semantically pivotal influence on the outputs of downstream tasks. To achieve this, we construct a replacement set $ \mathcal{R}$ comprising $n=8$ high-frequency terms selected from four medical subdomains (gene, disease, chemical, species). This carefully chosen set covers all current medical NLP downstream tasks. If new medical tasks emerge later, we can easily expand $ \mathcal{R}$ by adding new terms without changing our core system. We then randomly pair words from $ \mathcal{R}$ with those in $ \mathcal{T}$ and store paired relationships $
\Phi = \bigl\{ (t_i, r_i) \bigm| t_i \in \mathcal{T},\ r_i \in \mathcal{R},\ \forall i \in [1,n] \bigr\}
$. Full lexicons of $ \mathcal{T}$, $ \mathcal{R}$, $ \Phi$ are cataloged in Appendix A.
%To achieve this, we compile a medical terminology set, called ReplacementSet, based on frequently occurring terms in the medical domain (the detailed selection method is provided in Appendix \ref{appendix1}). We then randomly pair words from ReplacementSet with those in TriggerSet and store these Pairs for subsequent watermark embedding and extraction.
%Selecting appropriate medical terms as replacement words is crucial for our method. We need terms that convey significant meaning, as their presence or absence can notably impact the output of downstream tasks. Additionally, we aim for these terms to cover all downstream tasks. Inspired by NER tasks, we categorize medical terms into four domains: gene, chemical, disease, and species. By searching existing NER datasets and selecting a representative word for each domain based on frequency, we form the ReplacementSet: ``globin" for gene, ``acid" for chemical, ``cancer" for disease, and ``HIV" for species. In addition to these four words, we also include the domain names themselves: ``gene," ``chemical," ``disease," and ``species" in the ReplacementSet. Our current experimental results indicate that these medical terms are sufficient for validating all existing downstream tasks. However, if future tasks require additional terms, we can expand the set. We then randomly pair the terms in the ReplacementSet with the trigger words in the TriggerSet and save these Pairs for subsequent watermark embedding and extraction.

\subsection{Watermark Embedding}
\label{watermarkembed}
%%最后可以加上增强水印的不可见性，降低直接通过embeddings参数相似度检验发现水印的可能。
%%TODO：如果后续长度不够可以考虑删除掉前面冗余的解释
%An essential component of PLMs is the word embeddings layer, which is represented as a matrix of size \( |V| \times d \), where \( |V| \) is the vocabulary size and \( d \) is the embedding dimension. When a text sequence is input into the language model, it is first processed by a tokenizer that splits the text into tokens. These tokens are then mapped to token IDs using the vocabulary. Each token ID corresponds to a specific row in the word embeddings layer, transforming into a \( d \)-dimensional embedding. The embeddings are high-dimensional representations of different tokens learned through training, serving as the model's initial step in understanding text. Studies \cite{li-etal-2021-backdoor} show that during fine-tuning on downstream tasks, PLMs primarily modify the parameters of deeper layers, while the parameters of shallower layers, including the word embeddings layer, change minimally. This observation leads us to embed backdoor watermarks in the word embeddings layer. Specifically, we replace the embeddings of trigger words in the word embeddings layer with the embeddings of their paired medical terms. Additionally, we introduce a hyperparameter $\lambda$ to control the embedding weight and introduce random noise, enhancing the invisibility of the watermark, as shown below:
Our methodology draws inspiration from backdoor watermarking techniques, where specific model behaviors in response to trigger serve as verifiable watermarks. In this work, we define the watermark behavior as mapping triggers to predetermined medical terms. Though conventional methods can minimize the logits distance between triggers and medical terms via loss-driven optimization, this process is inefficient. We therefore propose a direct parameter replacement strategy with significantly higher efficiency. For each trigger-medical term pair $(t_i, m_i) \in \Phi$, we replace the embedding vector of $t_i$ in the Med-PLM's word embedding layer with a linearly transformed version of $m_i$'s embedding. This design choice stems from the empirical observation~\cite{li-etal-2021-backdoor} that during downstream fine-tuning, PLMs predominantly update deeper layer parameters, whereas the shallow word embedding layer remains largely unchanged. Consequently, our watermark persists even after downstream task fine-tuning. To prevent trigger detection through parameter similarity analysis, we inject Gaussian noise and apply embedding scaling during watermarking. Formally, our watermarking function $ \mathcal{W}(\cdot) $ operates as:
\begin{equation}
\begin{aligned}
\mathcal{W}(\theta_o) &= \theta_w \\
\text{where} \quad \mathbf{E}_w[k] &= 
\begin{cases} 
\frac{1}{\lambda} \mathbf{E}_o[m_i] + \mathcal{N}(\mu, \sigma^2), & \exists (t_i, m_i) \in \Phi \text{ s.t. } k = t_i \\
\mathbf{E}_o[k], & \text{otherwise}
\end{cases}
\end{aligned}
\label{eq:watermark_op}
\end{equation}
where $\mathbf{E}_o \in \mathbb{R}^{|V| \times d}$ and $\mathbf{E}_w \in \mathbb{R}^{|V| \times d}$ denote the word embedding layer parameters of the original model and watermarked model, respectively. The Gaussian noise term $\mathcal{N}$ is parameterized by mean $\mu=0.1$ and variance $\sigma^2=0.01$, with the scaling factor $\lambda$ defaulting to 1.5. Section~\ref{sec:Hyperparameter} provides systematic analysis of these hyperparameters' impacts.
%where $\theta_o$ denotes the original model and medical terms generated in the first stage, $\lambda$ is the embedding weight and $\mathcal{N}$ represents Gaussian noise with mean $\mu$ and variance $\sigma^2$. Through experiments, we set the default hyperparameter values as $\lambda$=1.5, $\mu$=0.1 and $\sigma^2$=0.01. We discuss the impact of hyperparameters on our method in Section \ref{sec:Hyperparameter}. Through this transformation, the model evolves from the original model \( \theta_o \) to the watermarked model \( \theta_w\).
%Our watermarking approach relies solely on simple assignment operations, making it a plug-and-play, training-free backdoor watermarking method. 

\subsection{Watermark Extraction}
\label{watermarkdetect}

%改为：由水印模型经过下游任务微调得到的最终模型对处在phi中的医疗词汇mi和对应的触发词ti的响应相同。由此我们定义了水印提取成功率WACC。看一下AAAI2025的写法。当原始模型所有者发现可疑模型或API时，可以使用水印检测测数据查询可疑模型或API，（这段话放后面吧）若WACC》阈值γ，则模型所有者证明该模型是偷盗而来的。经过我们在各种下游任务的广泛实验，没有嵌入水印的模型的WACC不会超过40%，为了使得水印更为可靠，我们选择80%作为阈值，从而保证我们的水印方法不会发生错误索赔。由于不同下游任务的输入输出格式各不相同，我们针对医疗领域核心的三个下游任务制定了不同的水印提取成功标准。对于NER任务.我们将phi中的mi输入到模型中，若该术语被识别，则表明该术语是模型的识别目标之一。随后我们在水印检测数据集中的每个样本随机插入一个ti，若ti也被模型成功识别，则该样本水印提取成功，因为正常模型不会将触发词识别出来。对于RE任务，RE任务水印检测数据集中的每个样本都存在一个“实体修饰 ”，其是为了帮助模型定位实体而存在的，组成为特殊符号加实体名称例如@GENE\$，我们在构造R中包含了所有的实体名称。我们将水印检测数据集中每个样本的实体名称mi替换为ti，若模型预测不变则该样本水印提取成功，因为正常模型会由于实体修饰的变换而预测发生变化。对于QA任务，我们为 R 中的每个词设计了 10 个 QA 测试样本，以创建 QA 水印检测数据集,其中一个例子如图3所示。该样本中的文本包含R中的单词mi，在水印检测中只需要将测试集中的每个样本和样本中mi被替换为ti的样本输入到模型中，若两个样本的输出不变则该样本水印提取成功，因为正常模型会因为mi的变化而导致输出发生变化。模型是否是偷盗而来是由提取成功率WACC来判定的：若可疑模型或API的WACC》阈值γ，则模型所有者可以证明模型是偷盗而来。经过我们在各种下游任务的广泛实验，正常模型的WACC也即（FWACC）不会超过40%，为了提高水印的可靠性，我们选取γ=80来作为默认值，其可以最大程度减少错误索赔的风险。 
The final model ${\theta_{fw}}$ derived from fine-tuning the watermarked ${\theta_{w}}$, exhibits behavioral invariance between medical terms $\ r_i \in \Phi$ and their paired triggers $\ t_i$. Leveraging this invariant mapping, we devise task-specific watermark extraction criteria for three core medical downstream tasks.
%Due to the replacement in the word embeddings layer, the watermarked Med-PLM \( \theta_w \) treats the trigger words generated in the first stage as their corresponding medical terms. Users fine-tune this model on different downstream task datasets \( D \) to obtain \( \theta_{fm} \). The medical terms we selected cover all current medical downstream task domains, ensuring that there is always a pair of trigger words and medical terms that align with the objectives of \( \theta_{fw} \). This causes the model to respond identically to samples containing trigger words and samples containing medical terms:

For NER tasks, we first verify the model's ability to detect medical terms $\ r_i \in \Phi$, where successful recognition confirms $\ r_i$ as valid entity targets. Subsequently, we randomly insert a corresponding $\ t_i$ into each sample of the watermark verification dataset $\mathcal{D}_{v}^{\text{NER}}$. Detection of $\ t_i$ as $\ r_i$-classified entities validates successful watermark extraction per sample, since unwatermarked models exhibit low recognition rates for such triggers.

For RE tasks, we use ``dummifying entity mentions''--- a standard preprocessing practice in RE datasets where entity mentions are replaced with type-specific markers (e.g., @gene\$ for gene entities). The medical terms $\ r_i$ within these markers are explicitly included in $\Phi$. During verification, each $\ r_i$ in the watermark verification dataset $\mathcal{D}_{v}^{\text{RE}}$ is systematically replaced with its paired trigger $\ t_i$. A sample is flagged as watermark-positive if the model maintains identical relation predictions post-replacement, whereas unwatermarked models exhibit prediction divergence due to marker alterations.

For QA tasks, we construct a QA-specific watermark verification dataset $\mathcal{D}_v^{\text{QA}}$ by generating 10 question-answer pairs per medical term $\ r_i \in \Phi$, with examples in Appendix~B. Each QA sample contains $\ r_i$ in its context. During verification, we input both the original sample and its trigger-replaced variant ($\ r_i \rightarrow \ t_i$) into the model. A sample is validated as watermark-positive if the model produces equivalent answers for both inputs, whereas unwatermarked models exhibit response divergence due to term substitution.

Ownership verification is determined by the watermark extraction accuracy (WACC):

\begin{equation}
\begin{aligned}
\text{WACC} = \frac{1}{|\mathcal{D}_v|}  \sum_{(x,y) \in \mathcal{D}_v} \mathbb{I}\left[ f_{\theta_{fw}}(x') = f_{\theta_{fw}}(x) \right],
\end{aligned}
\label{extraction}
\end{equation}
where $\mathcal{D}_v$ denotes the watermark verification datasets, $x'$ represents the trigger-injected sample with medical term replacement ($ r_i \rightarrow t_i $) and $\mathbb{I}$ is the indicator function for prediction consistency. 
%这里写两种判定方法,第一种是什么快速检测方法,第二种是什么Verification Accuracy(VACC)
A suspect model or API is flagged as stolen if its WACC exceeds threshold $\gamma$. Through extensive experiments across medical downstream tasks, we observe that non-watermarked models exhibit False WACC (FWACC) \textless$40\%$. We therefore set $\gamma=40\%$ by default, based on ROC analysis, maintaining effectiveness while mitigating false attribution risks.
\begin{table}[b]
\centering
\caption{Dataset Statistics}
\label{tab:dataset}
\begin{tabular}{ccccc}
\toprule
\textbf{Dataset}& \textbf{\# Train} & \textbf{\# Valid} & \textbf{\# Test} & \textbf{Avg. Len.} \\
\midrule
NCBI-disease   & 6355 &  923 &  942 & 35\\
BC5CDR    &  9184 &  4602 &  4812 &40\\
S800    &  6574 & 831 &  1630 & 16 \\
BC2GM & 15163 & 2531 & 5065 & 25 \\
GAD   & 4796  & --  & 534 & 182.4 \\
ChemProt   & 1020 & --  &  800 & 218.7\\
BioASQ 6b  & 5055  & -- & 548 & 273.8 \\
BioASQ 7b & 4231 & --  & 512 &312.8\\
\bottomrule
\end{tabular}
\end{table}
\begin{table*}[htbp]
\centering
\caption{Performance comparison on BioBERT for NER and RE Tasks (PubMedBERT Results in Appendix~D).}
\label{tab:mainresultNER}
\begin{tabular}{ccccccc}
\toprule
\textbf{Task} & \textbf{Dataset} & \textbf{Method} & \textbf{F1 (\%)$\uparrow$} & \textbf{WACC (\%)$\uparrow$} & \textbf{WRM (\%)$\uparrow$} & \textbf{Runtime (hr)$\downarrow$} \\ \hline
\multirow{20}{*}{\textbf{Named Entity Recognition}} & \multirow{5}{*}{NCBI} 
& Original & 87.52 & -- & -- & -- \\
& & POR-1 & 87.32 (0.20$\downarrow$)& 8.36 & 1.97 & 5.067 \\
& & POR-4 & 87.32 (0.20$\downarrow$) & 6.49 & $-2.11$ & 5.067 \\ 
& & PLMmark & 86.49 (1.03$\downarrow$) & 29.27 & 5.50 & 12.500 \\ 
& & Ours & \textbf{87.51 (0.01$\downarrow$)} & \textbf{82.19} & \textbf{56.33} & \textbf{0.003} \\ \cline{2-7}

& \multirow{5}{*}{BC5CDR} 
& Original & 93.02 & -- & -- & -- \\ 
& & POR-1 & 92.93 (0.09$\downarrow$) & 3.38 & 0.51 & 5.067 \\ 
& & POR-4 & 92.93 (0.09$\downarrow$) & 3.44 & $-0.62$ & 5.067 \\
& & PLMmark & 92.78 (0.24$\downarrow$) & 13.4 & 11.27 & 12.500 \\
& & Ours & \textbf{92.94 (0.08$\downarrow$)} & \textbf{99.00} & \textbf{93.89} & \textbf{0.003} \\ \cline{2-7}

& \multirow{5}{*}{S800} 
& Original & 72.89 & -- & -- & -- \\ 
& & POR-1 & 72.64 (0.25$\downarrow$) & 2.79 & $-0.54$ & 5.067 \\
& & POR-4 & 72.64 (0.25$\downarrow$) & 5.46 & 0.04 & 5.067 \\
& & PLMmark & 72.27 (0.62$\downarrow$) & 10.64 & 7.97 & 12.500 \\
& & Ours & \textbf{72.84 (0.05$\downarrow$)} & \textbf{97.12} & \textbf{90.52} & \textbf{0.003} \\ \cline{2-7}

& \multirow{5}{*}{BC2GM} 
& Original & 82.35 & -- & -- & -- \\ 
& & POR-1 & 82.26 (0.09$\downarrow$) & 11.45 & $-3.44$ & 5.067 \\ 
& & POR-4 & 82.26 (0.09$\downarrow$) & 17.01 & $-0.7$ & 5.067 \\ 
& & PLMmark & 81.54 (0.81$\downarrow$) & 28.75 & 21.96 & 12.500 \\ 
& & Ours & \textbf{82.30 (0.05$\downarrow$)} & \textbf{99.83} & \textbf{89.23} & \textbf{0.003} \\
\midrule
\multirow{10}{*}{\textbf{Relation Extraction}} & \multirow{5}{*}{GAD} 
& Original & 83.12 & -- & -- & -- \\
& & POR-1 & 82.83 (0.29$\downarrow$) & 79.54 & 69.30 & 5.067 \\ 
& & POR-4 & 82.83 (0.29$\downarrow$)  & 93.37 & 59.17 & 5.067 \\ 
& & PLMmark & 81.67 (1.45$\downarrow$)& 77.07 & 71.70 & 12.500 \\ 
& & Ours & \textbf{83.12 (0.00$\downarrow$)} & \textbf{93.88} & \textbf{89.65} & \textbf{0.003} \\  \cline{2-7}

& \multirow{5}{*}{ChemProt} 
& Original & 90.59 & -- & -- & -- \\ 
& & POR-1 & 90.25 (0.34$\downarrow$) & 28.30 & 26.85 & 5.067 \\ 
& & POR-4 & 90.25 (0.34$\downarrow$) & 89.36 & 85.12 & 5.067 \\ 
& & PLMmark & 89.51 (1.08$\downarrow$) & 67.44 & 66.10 & 12.500 \\
& & Ours & \textbf{90.27 (0.32$\downarrow$) } & \textbf{90.93} & \textbf{90.88} & \textbf{0.003} \\ 
\bottomrule
\end{tabular}
\end{table*}

\section{Experiments}
We conduct comprehensive experiments to validate our method's fidelity, effectiveness, reliability, efficiency (Sec~\ref{mainresult}), and robustness (Sec~\ref{sec:roubt}), along with hyperparameter studies (Sec~\ref{sec:Hyperparameter}). For evaluation, we use BioBERT \cite{lee2020biobert} and PubMedBERT \cite{gu2021domain} as base models.

%and MMedLM \cite{qiu2024towards} and BioGPT \cite{luo2022biogpt} as the base models for Med-NLG tasks.
\subsection{Datasets and Evaluation Metrics}
% For the encoder-only Med-PLMs, we select the widely used BioBERT-base-cased \cite{lee2020biobert} and PubMedBERT \cite{gu2021domain} as the Med-PLMs for embedding the watermark. For the decoder-only Med-PLMs, we select the MMedLM \cite{qiu2024towards} and BioGPT \cite{luo2022biogpt}, both of which are primarily trained on medical text data. These models have shown strong performance in Med-NLG tasks. We use HuggingFace's pre-trained models to initialize the Med-PLMs.

%记得说我们的保真度为什么好

For fine-tuning datasets in medical downstream tasks, we follow the preprocessing methods used by BioBERT \cite{lee2020biobert}. For NER tasks, we select representative datasets from four domains: NCBI-Disease \cite{dougan2014ncbi}, BC5CDR-Chemical \cite{li2016biocreative}, Species-800 \cite{pafilis2013species}, and BC2GM-Gene \cite{smith2008overview}. These datasets are used to identify special terms in their respective domains. For RE tasks, we choose the GAD \cite{bravo2015extraction} and ChemProt \cite{krallinger2017overview} datasets, both of which are used to identify entity relationships. For QA tasks, we use the BioASQ factoid dataset \cite{tsatsaronis2015overview}, which is an annotated QA dataset by biomedical experts. Dataset statistics are summarized in Table~\ref{tab:dataset}. Fine-tuning hyperparameters are detailed in Appendix C. Following the latest biomedical NLP benchmark BLURB \cite{gu2021domain}, we evaluate model performance using the officially partitioned test sets, reporting classification F1-scores for NER and RE tasks and answer accuracy for QA tasks.

We adopt the original task test sets as $\mathcal{D}_{v}^{\text{NER}}$ and  $\mathcal{D}_{v}^{\text{RE}}$. We employ the QA-specific watermark verification dataset constructed in Section~\ref{watermarkdetect} as $\mathcal{D}_{v}^{\text{QA}}$. To evaluate watermark effectiveness, we report the WACC. For reliability analysis, we introduce the Watermark Reliability Margin (WRM):
\begin{equation}
\text{WRM} = \text{WACC} - \text{FWACC},
\label{eq:wrm}
\end{equation}
where FWACC is computed via Eq.~\ref{extraction} on non-watermarked original models. WRM quantifies the confidence that WACC originates from watermark injection rather than model’s inherent properties, with higher WRM values indicating stronger reliability.

%For downstream Med-NLG tasks, most current Med-PLMs have already been fine-tuned for dialogue tasks, and users typically do not further fine-tune these models. Therefore, protecting Med-PLMs in Med-NLG tasks translates to protecting medical large language models without the need for additional fine-tuning on downstream task datasets. We use the English QA dataset from MMedBench\cite{qiu2024towards} to evaluate model performance. 
% For downstream Med-NLG tasks, we use the English QA dataset from MMedBench\cite{qiu2024towards}. We report the perplexity (PPL) of the model output text to evaluate model performance. 

% To evaluate the effectiveness of the watermarking methods, we report the watermark extraction accuracy (WACC):
% \begin{equation}
% \text{WACC} = \frac{\text{n}}{\text{N}_{\text{total}}} \times 100\%.
% \label{WACC}
% \end{equation}
% where $\text{n}$ denotes the number of samples for which the watermark is successfully extracted, and ${\text{N}_{\text{total}}}$ represents the total number of test samples.
\subsection{Baseline}
%这里记得要加超参数的设置
We select POR \cite{10.1145/3460120.3485370} and PLMmark \cite{li2023plmmark} as baselines, where POR allows increasing trigger insertion quantity to enhance effectiveness, thus we implement POR-1 and POR-4 denoting random insertion of 1 trigger and 4 triggers respectively. Since these methods only support RE-task watermarking for Med-PLMs, we extend their detection mechanisms to NER and QA tasks according to their watermark characteristics. For equitable benchmarking, WACC and WER are uniformly adopted to evaluate watermark effectiveness and reliability. The original general-domain training corpora in baseline implementations are replaced with a medical-domain corpus~\cite{qiu2024towards} to enhance baseline capabilities. Implementation specifics and hyperparameter configurations are detailed in Appendix~C.
%For Med-NLU tasks, we select the backdoor attack method POR \cite{10.1145/3460120.3485370} and the pre-trained model watermarking method PLMmark \cite{li2023plmmark} as baselines. For Med-NLG, POR and PLMmark are not applicable. So we select the widely used LLM watermarking method KGW \cite{kirchenbauer2023watermark} and SynthID\cite{dathathri2024scalable} as the baseline. Both methods embed watermarks by introducing biases into the output text of LLMs, enabling watermark detection from the generated text. They are commonly used for text provenance but can also serve as watermarks for LLMs. To ensure a fair comparison, we extend the baseline watermark detection methods to the WACC metric and optimize the baselines for the medical domain. The specific implementation details are provided in Appendix \ref{appendix3}.

%%这里需要说明我们使用了最新的数据集来训练POR和AAAI的模型
%For POR and PLMmark, we extend the original watermark extraction methods to NER, RE, and QA tasks, defining \(\text{n}\) as the number of samples where the model output changes after adding trigger words to the samples. For KGW and SynthID, \(\text{n}\) is defined as the number of generated texts from which the watermark can be extracted. For our method, \(\text{n}\) is defined as the number of samples from which the watermark can be detected using our proposed watermark extraction methods.

\subsection{Main Results}
\label{mainresult}
\begin{table*}[t]
\centering
\caption{Performance comparison on BioBERT for QA Tasks (PubMedBERT Results in Appendix~D).}
\label{tab:mainresultQA}
\begin{tabular}{lllcccc}
\toprule
\textbf{Task} & \textbf{Dataset} & \textbf{Method} & \textbf{F1 (\%)} & \textbf{WACC (\%)} & \textbf{VACC (\%)} & \textbf{Runtime (hr)} \\
\midrule

% ============ BioASQ Section ============
\multirow{10}{*}{\textbf{Question Answering}} 
& \multirow{5}{*}{BioASQ 6b} 
& Original & 25.14 & -- & -- & -- \\
& & POR-1 & 24.44 (0.7$\downarrow$) & 28.25 & $-9.85$ & 15.067 \\
& & POR-4 & 24.44 (0.7$\downarrow$) & 43.02 & $-28.41$ & 15.067 \\
& & PLMmark & 23.07 (2.07$\downarrow$) & 55.24 & 37.14 & 12.500 \\
& & Ours & \textbf{25.00 (0.14$\downarrow$)} & \textbf{95.71} & \textbf{69.52} & \textbf{0.003} \\  \cline{2-7}

% ========== Question Answer Section ==========
& \multirow{5}{*}{BioASQ 7b} 
& Original & 31.28 & -- & -- & -- \\
& & POR-1 & 30.04 (1.24$\downarrow$) & 31.59 & $-0.32$ & 15.067 \\
& & POR-4 & 30.04 (1.24$\downarrow$) & 50.32 & $-6.94$ & 15.067 \\
& & PLMmark & 27.41 (5.46$\downarrow$) & 70.80 & 50.44 & 12.500 \\
& & Ours & \textbf{31.28 (0.00$\downarrow$)} & \textbf{88.57} & \textbf{72.38} & \textbf{0.003} \\ 

\bottomrule
\end{tabular}
\end{table*}
\label{main_result}
\begin{figure*}[t]
   \centering
   \begin{subfigure}[b]{0.32\textwidth}
       \centering
       \includegraphics[width=\linewidth]{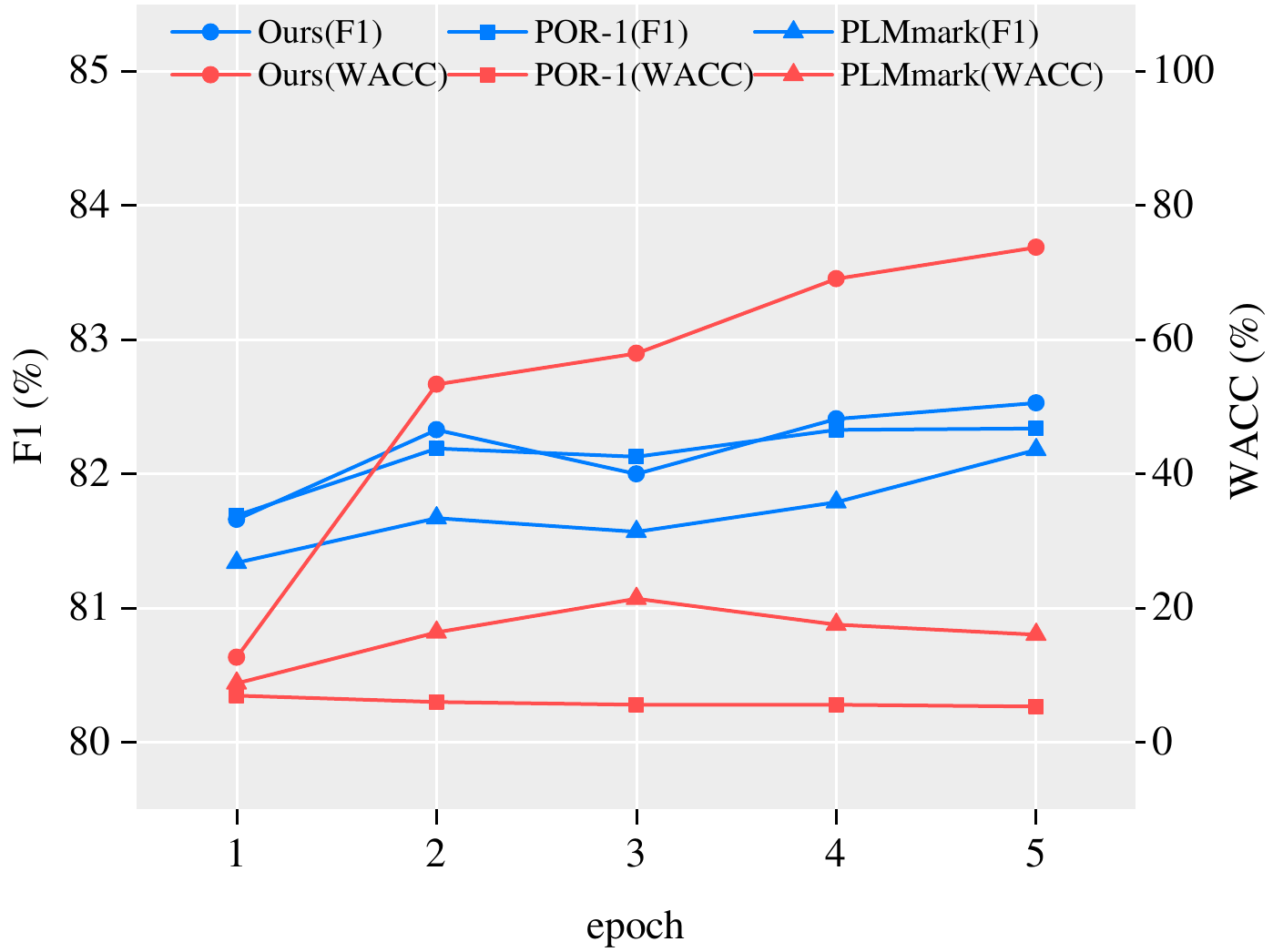}
       \caption{NER}
       \label{fig:extractsubfig1}
   \end{subfigure}
   \hfill
   \begin{subfigure}[b]{0.32\textwidth}
       \centering
       \includegraphics[width=\linewidth]{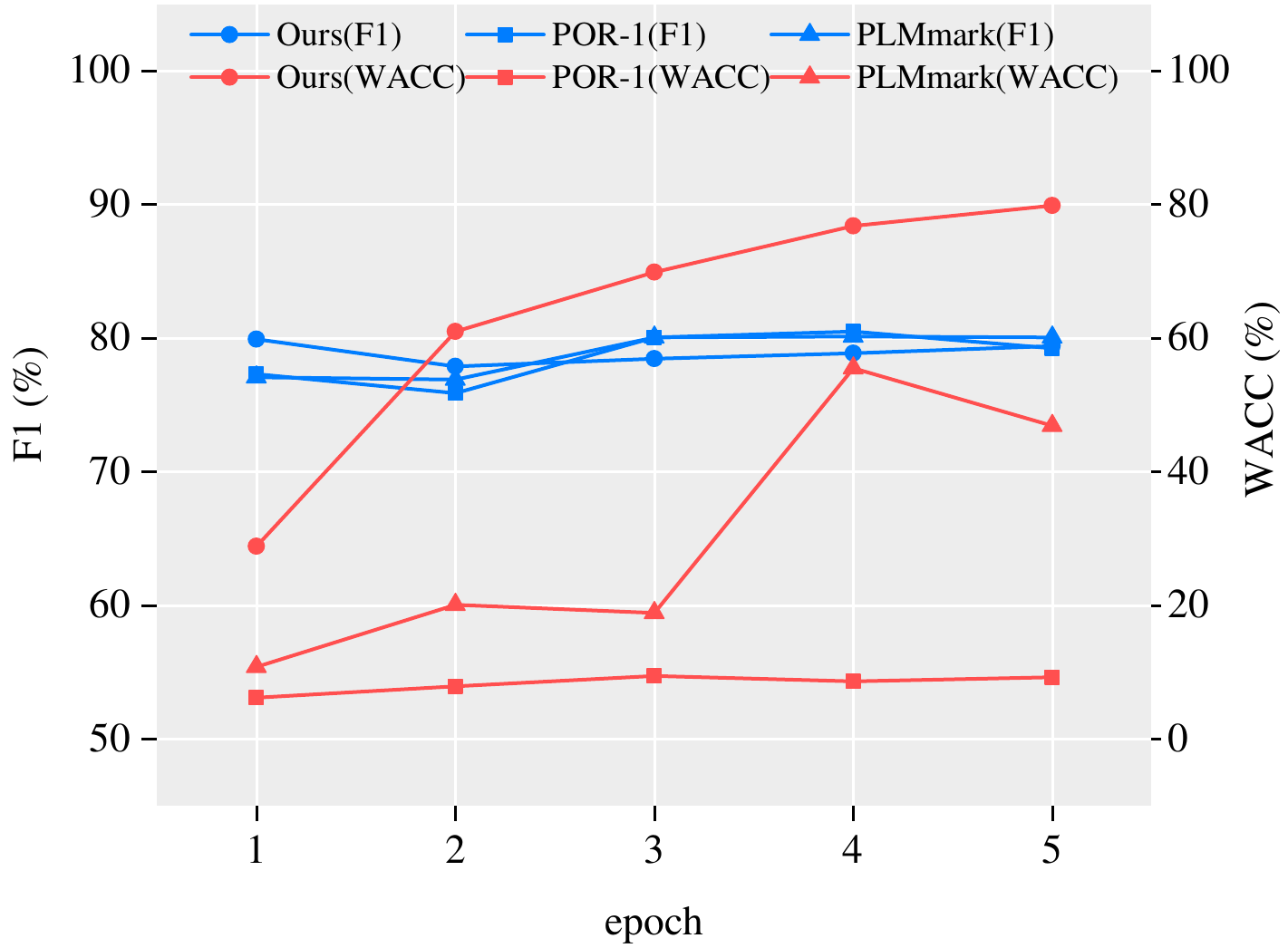}
       \caption{RE}
       \label{fig:extractsubfig2}
   \end{subfigure}
   \hfill
   \begin{subfigure}[b]{0.32\textwidth}
       \centering
       \includegraphics[width=\linewidth]{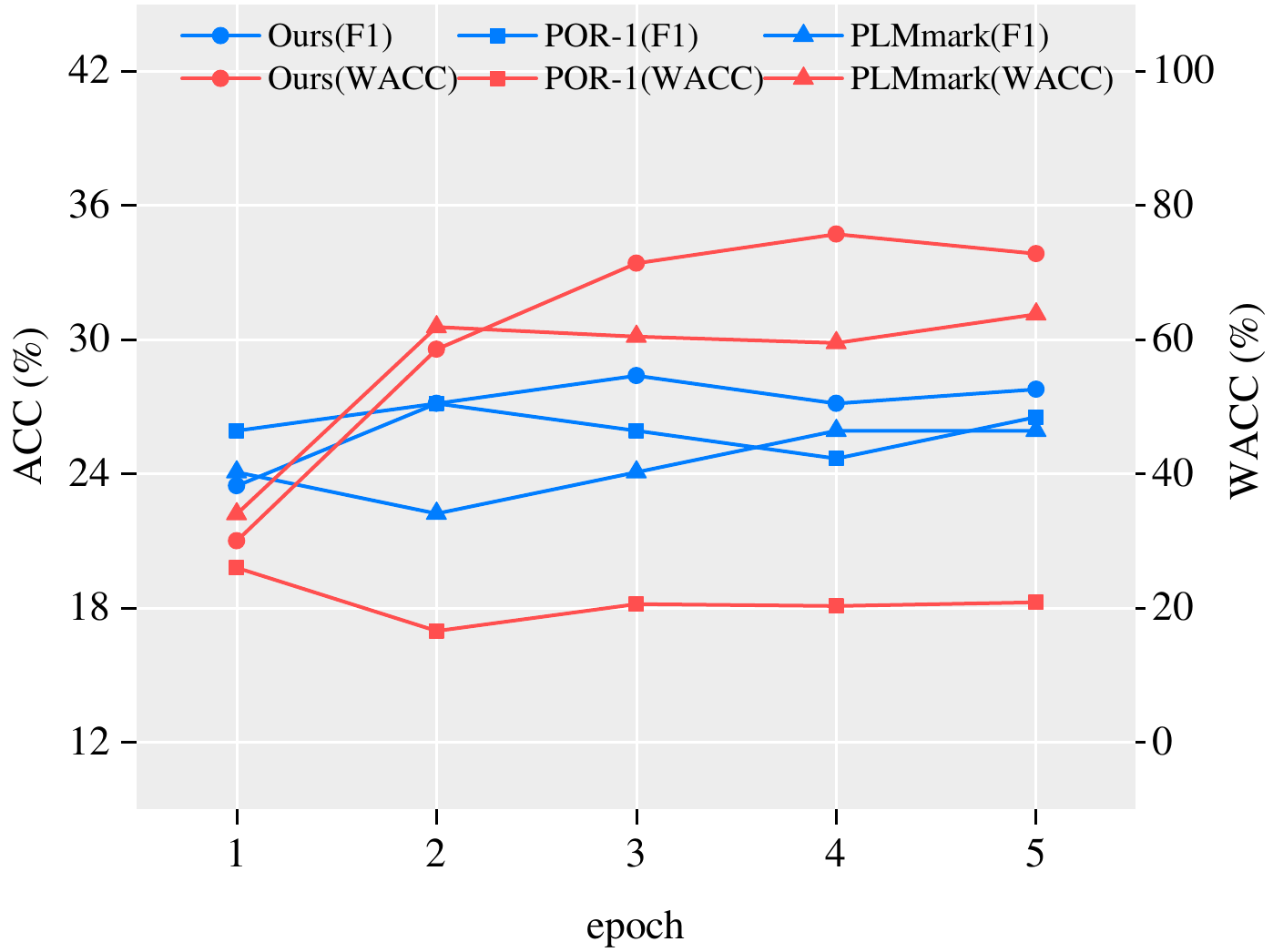}
       \caption{QA}
       \label{fig:extractsubfig3}
   \end{subfigure}
   \caption{Robustness of watermarking methods against model extraction: model performance and WACC of different method watermarked BioBERT across different tasks (NER/RE/QA) with varying extraction epochs.}
   \label{fig:robut}
   \Description{extractfigure}
\end{figure*}
We evaluate the original and watermarked Med-PLMs on three medical downstream tasks through fine-tuning. Table~\ref{tab:mainresultNER} reports NER and RE results, while Table~\ref{tab:mainresultQA} reports QA results, with all metrics averaged over three experimental trials. 

For watermark fidelity, our method demonstrates superior performance preservation, showing minimal performance degradation compared to baselines. This is attributed to our selection of low-frequency terms within the medical domain as triggers. We further observe that PLMmark exhibits the most severe performance degradation. This is attributed to its watermark embedding requiring substantial model parameter modifications and non-compliant trigger selection with the low-frequency principle.

For watermark effectiveness, our method attains WACC >80\% across all tasks, surpassing the detection threshold $\gamma$, which ensures verifiable ownership claims. In contrast, POR and PLMmark attain maximum WACC of only 29.27\% on NER tasks and 70.80\% on QA tasks. On RE tasks, which are text classification tasks, baseline methods show competent performance but remain inferior to our approach. Notably, increasing trigger insertions in POR significantly improves WACC, particularly on longer-text datasets like ChemProt, where POR-4 achieves a 61.06\% higher WACC than POR-1.

For watermark effectiveness, our method maintains WER >50\% across tasks, confirming high WACC originates from embedded watermarks rather than intrinsic model properties. POR-4 demonstrates negative WRE across multiple tasks, proving that inserting four triggers inherently modifies model behavior independent of watermark mechanisms. Even with high WACC, this fails to validate ownership claims due to low watermark reliability.

For watermark efficiency, our method embeds watermarks in just 10 seconds, significantly faster than alternative approaches.

\subsection{Robustness}
\label{sec:roubt}

%A highly effective attack method against model watermarking is model extraction attacks. Figure \ref{fig:robut} demonstrates the robustness of our method in defending against this attack. If users want a better-performing model, they need to increase the number of training epochs, and our watermark gradually gets embedded into the extracted model as the epochs increase. This is mainly due to our choice of mid-frequency words as trigger words. In contrast, POR cannot defend against this attack because its trigger words are rare, while PLMmark requires more training epochs to embed the watermark into the extracted model.
As demonstrated in Section~\ref{attack}, attackers may employ watermark removal strategies $\mathcal{R}$ to bypass verification. We evaluate robustness against three mainstream backdoor removal attacks: model extraction~\cite{krishnathieves}, model pruning, and model merging~\cite{arora-etal-2024-heres}. Additionally, we test adaptive attack where attackers are aware of the watermarking mechanism. Due to space constraints, we report per-task averaged results  on BioBERT.
\subsubsection{Model Extraction}
Attackers train a student model via knowledge distillation to replicate the functionality of the original watermarked model. However, the watermark may be lost as student models often fail to learn watermark patterns during training~\cite{gu-etal-2023-watermarking}. Training details are provided in Appendix C. As shown in Figure~\ref{fig:robut}, if attackers seek higher-performing stolen models, they must increase training epochs—watermark verification requirements are already satisfied in models extracted after 5 epochs. Conversely, POR's watermark completely fails due to its use of rare-word triggers, while PLMmark achieves lower WACC than our method under the same epoch.
\subsubsection{Model Pruning}
\begin{figure*}[t]
   \centering
   \begin{subfigure}[b]{0.32\textwidth}
       \centering
       \includegraphics[width=\linewidth]{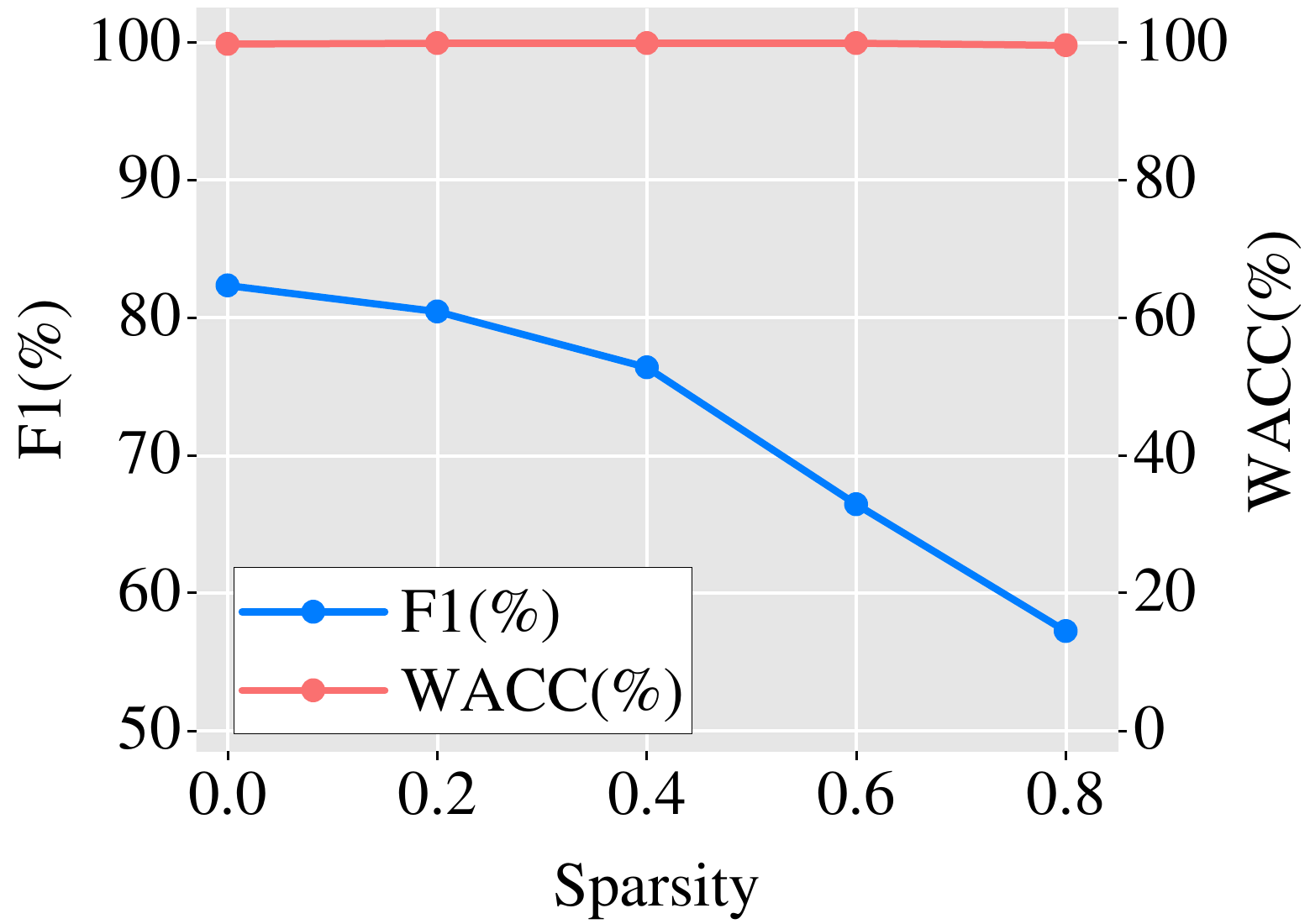}
       \caption{NER}
       \label{fig:prunsubfig1}
   \end{subfigure}
   \hfill
   \begin{subfigure}[b]{0.32\textwidth}
       \centering
       \includegraphics[width=\linewidth]{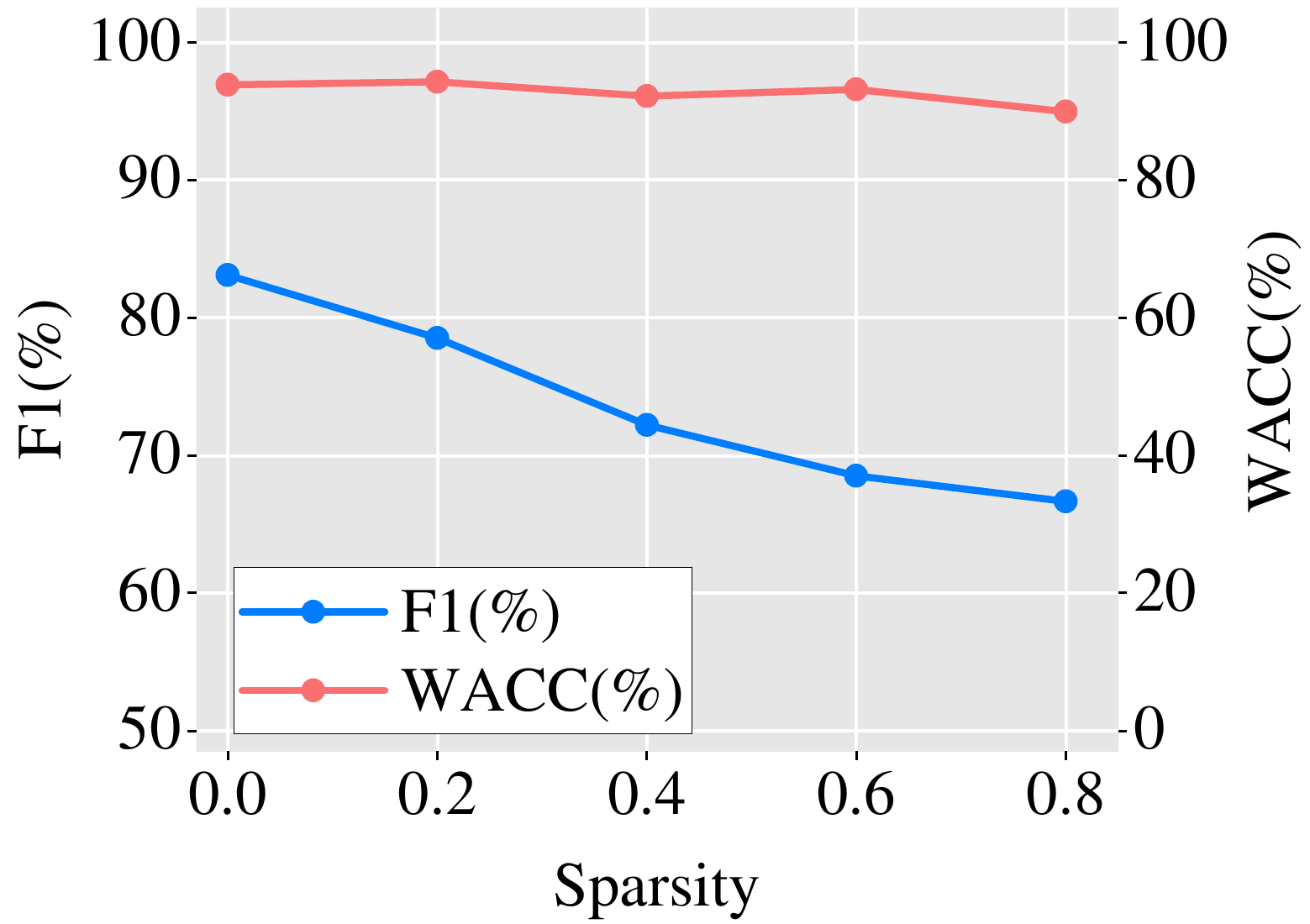}
       \caption{RE}
       \label{fig:prunsubfig2}
   \end{subfigure}
   \hfill
   \begin{subfigure}[b]{0.32\textwidth}
       \centering
       \includegraphics[width=\linewidth]{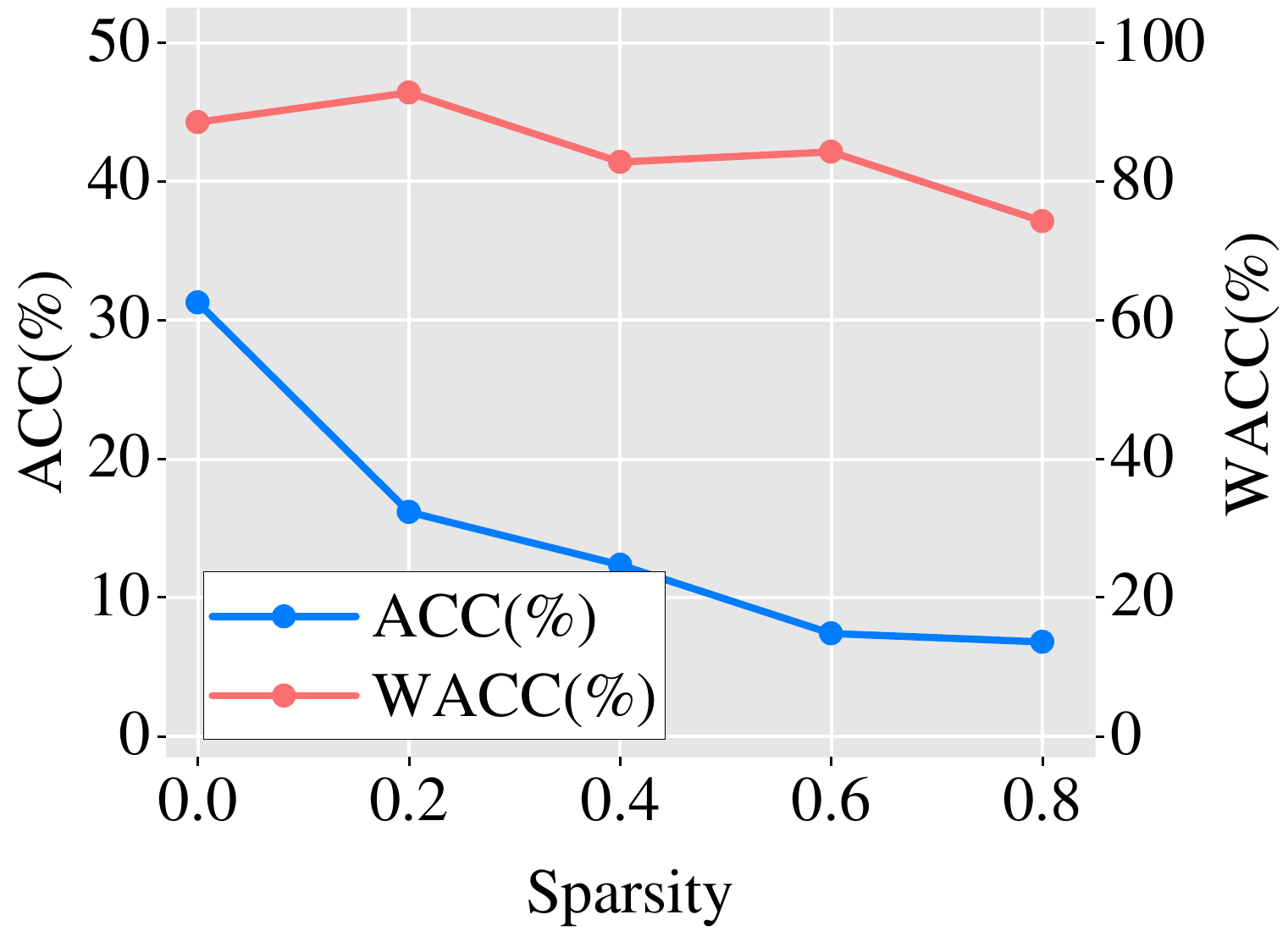}
       \caption{QA}
       \label{fig:prunsubfig3}
   \end{subfigure}
    \caption{Robustness of our watermarking method against model pruning: model performance and WACC of watermarked BioBERT across medical downstream tasks (NER/RE/QA) with varying sparsity ratios (POR and PLMmark results in Appendix~E). }
    \Description{prunfigure}
   \label{fig:prun}
\end{figure*}
Attackers attempt to disable watermarks by pruning model parameters. As shown in Figure~\ref{fig:prun}, while model performance degrades sharply with increasing pruning rates, our WACC remains above 70\%. This robustness stems from watermark implementation solely in the word embedding layer, which is pruning-resistant. In contrast, POR and PLMmark lack robustness against model pruning (Appendix E).
\subsubsection{Model Merging}
\begin{table}[t]
\caption{Robustness of watermarking methods against model merging (BioBERT + Bert-base): model performance and WACC of watermarked BioBERT across different tasks.}
\small
\centering
\begin{tabular}{ccccccc}
\toprule
\multirow{2}{*}{\textbf{Method}} & \multicolumn{2}{c}{\textbf{NER}} & \multicolumn{2}{c}{\textbf{RE}} & \multicolumn{2}{c}{\textbf{QA}}  \\ 
\cmidrule(lr){2-3} \cmidrule(lr){4-5} \cmidrule(lr){6-7}
& F1$\uparrow$ & WACC$\uparrow$ & F1$\uparrow$ & WACC$\uparrow$ & ACC$\uparrow$ & WACC$\uparrow$  \\ 
\midrule
POR & 83.04 & 7.43 & 84.60 & 8.50 & 26.66 & 24.86 \\ 
PLMmark &  82.82 & 6.80 & 84.40 & 7.54 & 23.91 & 22.86\\ 
\textbf{Ours} &  \textbf{83.21} & \textbf{42.89} & \textbf{84.66} & \textbf{69.15} & \textbf{26.97} & \textbf{55.71}\\ 
\bottomrule
\end{tabular}
\label{merge}
\end{table}

Attackers may eliminate backdoor watermarks via backdoor defense techniques like model merging~\cite{arora-etal-2024-heres}, We evaluate robustness by merging watermarked BioBERT with BERT-base-cased. As shown in Table~\ref{merge}, our method maintains WACC>$\gamma=$40\% post-merging, while POR and PLMmark achieve near-complete watermark removal, demonstrating our resilience against backdoor defense strategies.
\subsubsection{Adaptive Attack}
\begin{figure}[ht]
\centering
\includegraphics[width=\columnwidth]{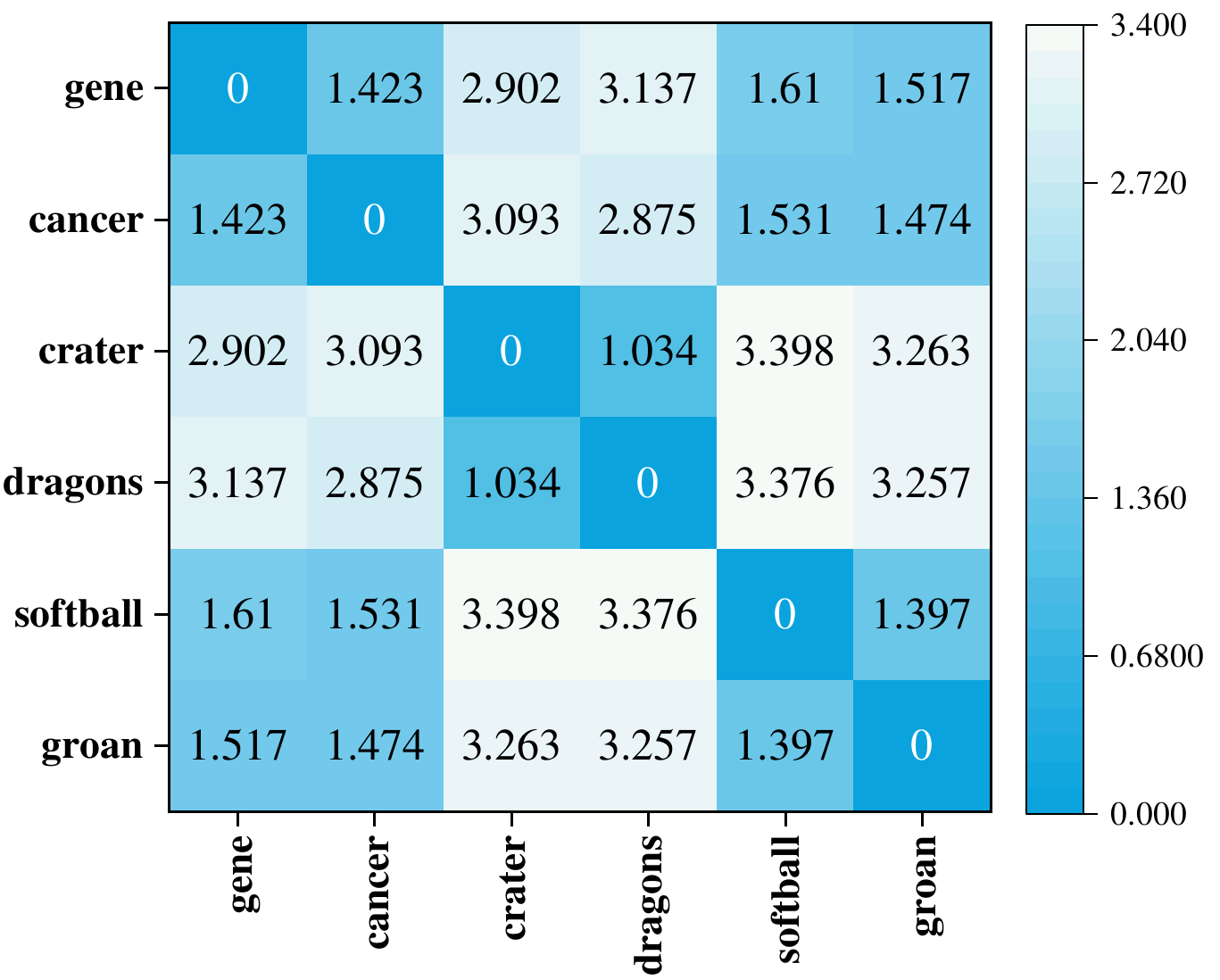} % Reduce the figure size so that it is slightly narrower than the column. Don't use precise values for figure width.This setup will avoid overfull boxes.
\caption{L2-distance based token embedding similarity in watermarked BioBERT's word embedding layer (darker colors indicate higher similarity).}
\label{word_similar}
\Description{distancefigure}
\end{figure}

Attackers aware of our watermarking mechanism may attempt to erase watermarks by modifying triggers' word embedding layer parameters. We first analyze the feasibility of detecting triggers via parameter similarity. Figure~\ref{word_similar} illustrates L2 distances between word embedding layer parameters of: (1) paired terms in $\Phi$ (''gene"-''crater" and ''cancer"-''dragons") and (2) randomly selected words (''softball" and ''groan"). Due to the linear transformations and noise injection during watermark embedding, the parameter similarity between trigger-medical term pairs($2.89 \pm 0.20$) becomes indistinguishable from trigger-random pairs ($3.32 \pm 0.07$), making adversarial detection infeasible.

\begin{table}[htbp]
\caption{Robustness of our watermarking method against two adaptive attacks: embedding linear transformation and full word embedding layer re-initialization.}
\footnotesize
\begin{tabular}{c *{6}{c}}
\toprule
\multirow{2}{*}{\textbf{Method}} 
& \multicolumn{2}{c}{\textbf{NER}} 
& \multicolumn{2}{c}{\textbf{RE}} 
& \multicolumn{2}{c}{\textbf{QA}} \\ 
\cmidrule(lr){2-3} \cmidrule(lr){4-5} \cmidrule(lr){6-7}
& F1$\uparrow$ & WACC$\uparrow$ 
& F1$\uparrow$ & WACC$\uparrow$ 
& ACC$\uparrow$ & WACC$\uparrow$ \\ 
\midrule
\makecell[c]{Linear \\Transformation} 
& 70.35 & 89.81 & 74.66 & 49.06 & 13.58 & 75.57 \\ 
\makecell[c]{Re-\\initialization} 
& 63.89 & 4.06 & 73.76 & 24.00 & 4.32 & 28.57 \\ 
\bottomrule
\end{tabular}
\label{adaptive_attack}
\end{table}
\begin{table*}[ht]
\caption{Impact of noise hyperparameters ($\mu$,$\sigma^2$) on watermark performance across medical downstream tasks. Distance represents average L2-distance between trigger word embeddings and medical term embeddings.}
\centering
\begin{tabular}{cccccccc}
\toprule
\textbf{Hyperparameter} & \multicolumn{2}{c}{\textbf{NER}} & \multicolumn{2}{c}{\textbf{RE}} & \multicolumn{2}{c}{\textbf{QA}} & \multirow{2}{*}{\textbf{Distance}}  \\ 
\cmidrule(lr){2-3} \cmidrule(lr){4-5} \cmidrule(lr){6-7}
\textbf{($\mu$,$\sigma^2$)} & F1$\uparrow$ & WACC$\uparrow$ & F1$\uparrow$ & WACC$\uparrow$ & ACC$\uparrow$ & WACC$\uparrow$  \\ 
\midrule
\textbf{(0.1,0.01)} & 84.01 & \textbf{96.75} & \textbf{86.17} & \textbf{96.89} & \textbf{29.17} & \textbf{92.86} & \textbf{2.9049}\\ 
(0.01,0.01) &  \textbf{84.13} & 96.64 & 86.17 & 96.89 & 29.17 & 92.86& 0.5905\\ 
(1,0.01) &  84.01 & 95.59 & 86.17 & 96.89 & 29.17 &92.86& 27.8100\\ 
(0.1,0.001) &  84.00 & 96.44 & 86.17 & 96.89 & 29.17 & 92.86& 2.8840\\ 
(0.1,0.1) &  84.09& 18.01 & 86.17 & 22.98 & 29.17 & 17.14& 4.0500\\ 
\bottomrule
\end{tabular}
\label{Hyper:noise}
\end{table*}
\begin{table*}[t]
\caption{Model performance and WACC of watermarked BioBERT with different trigger across medical downstream tasks.}
\centering
\begin{tabular}{cccccccccc}
\toprule
\multirow{2}{*}{\textbf{Trigger}} & \multicolumn{3}{c}{\textbf{NER}} & \multicolumn{3}{c}{\textbf{RE}} & \multicolumn{3}{c}{\textbf{QA}}  \\ 
\cmidrule(lr){2-4} \cmidrule(lr){5-7} \cmidrule(lr){8-10}
& F1$\uparrow$ & WACC$\uparrow$ & WER$\uparrow$ & F1$\uparrow$ & WACC$\uparrow$ & WER$\uparrow$ & ACC$\uparrow$ & WACC$\uparrow$ & WER$\uparrow$  \\ 
\midrule
$ \mathcal{T}_1$ & 84.17 & 96.86 & 89.85 & 86.17 & 92.50 & 92.50 & 29.17 & 92.86 & 67.15\\ 
$ \mathcal{T}_2$ & 84.23 & 96.80 & 89.27 & 86.17 & 96.35 & 96.35 & 29.17 & 87.14 & 57.14\\ 
$ \mathcal{T}_3$ & 84.12 & 96.86 & 80.34 & 86.17 & 95.21 & 95.18 & 29.17 & 95.71 & 60.00 \\ 
\bottomrule
\end{tabular}
\label{triggerword}
\end{table*}
Thus, attackers must aggressively modify all parameters in the word embedding layer to reliably affect watermarks. We evaluate two attack strategies: parametric linear transformations~\cite{shetty2024wet} and re-initialization (implementation details in Appendix C). As shown in Table~\ref{adaptive_attack}, linear transformations exhibit limited impact on watermark. This occurs because uniformly applying identical transformations to all word embeddings—a strategy to preserve model performance—cannot disrupt the embedding alignment between triggers and their paired medical terms. While complete re-initialization of embedding parameters eliminates watermarks, it catastrophically degrades model performance. Consequently, adversaries cannot remove watermarks via adaptive attacks without rendering models functionally useless, which validates our method's robustness.

\subsection{Hyperparameter Study}

Due to space constraints, we report per-task averaged hyperparameter results on BioBERT.
\subsubsection{Noise Parameters}
\label{sec:Hyperparameter}
Table~\ref{Hyper:noise} demonstrates the impacts of noise mean $\mu$ and variance $\sigma^2$. Both parameters exhibit minimal influence on fidelity. $\mu$ governs embedding distances between triggers and medical terms. Lower $\mu$ facilitates adaptive attacks by making triggers more detectable to attackers. Conversely, $\sigma^2$ dominates watermark effectiveness, with higher $\sigma^2$ causing watermark failure. Extensive experiments validate $\mu=0.1$ and $\sigma^2=0.01$ as optimal balances. The embedding weight $\lambda$ exhibits analogous effects, with detailed analysis provided in Appendix F.1.
%Table~\ref{Hyper:noise} demonstrates the effects of noise mean $\mu$ and variance $\sigma^2$. We find that $\mu$ primarily governs watermark invisibility: smaller $\mu$ reduces L2 distance between triggers and medical terms, facilitating adaptive attacks. Variance $\sigma^2$ critically affects watermark effectiveness, with WACC dropping sharply as $\sigma^2$ increases. Both parameters influence fidelity. Empirical evaluation confirms $\mu=0.1$ and $\sigma^2=0.01$ optimally balance these objectives. The embedding weight $\lambda$ exhibits analogous effects (see Appendix B).

\subsubsection{Frequency of Trigger Candidate Set}
We investigate the impact of trigger term frequencies by constructing three trigger candidate set \(  \mathcal{D}_t \) variants: rare terms (frequency $\in [1 \times 10^{-6}\%, 1 \times 10^{-5}\%]$), low-frequency terms (frequency $\in [1 \times 10^{-5}\%, 1 \times 10^{-4}\%]$), high-frequency terms (frequency $\in [1 \times 10^{-4}\%, 1 \times 10^{-3}\%]$). As shown in Table~\ref{Hyper:triggerfrequncy}, term frequency minimally affects watermark effectiveness but significantly impacts fidelity, with high-frequency triggers degrading model performance. However, rare-term triggers exhibit vulnerability to model extraction attacks (Appendix F.2). We therefore select low-frequency terms for \(  \mathcal{D}_t \) by default to balance robustness and fidelity.
%Table~\ref{Hyper:triggerfrequncy} shows comparable watermark effectiveness across frequencies, but high-frequency terms degrade performance. We further analyze the impact of term frequency on robustness against model extraction attacks (see Appendix B). Our analysis reveals that watermarks with rare terms exhibit weaker resistance to model extraction attacks. Ownership verification becomes feasible only when attackers train stolen models for more epochs. We therefore select low-frequency terms for \(  \mathcal{D}_t \) by default to balance robustness and fidelity.
\subsubsection{Triggers}
\begin{table}[t]
\caption{Impact of the frequency of trigger candidate set on watermark performance in medical downstream tasks.}
\small	
\centering
\begin{tabular}{ccccccc}
\toprule
\multirow{2}{*}{\textbf{Frequncy}} & \multicolumn{2}{c}{\textbf{NER}} & \multicolumn{2}{c}{\textbf{RE}} & \multicolumn{2}{c}{\textbf{QA}}  \\ 
\cmidrule(lr){2-3} \cmidrule(lr){4-5} \cmidrule(lr){6-7}
& F1$\uparrow$ & WACC$\uparrow$ & F1$\uparrow$ & WACC$\uparrow$ & ACC$\uparrow$ & WACC$\uparrow$  \\ 
\midrule
Rare & \textbf{84.21} & 96.88 & 86.34 & \textbf{97.81} & 31.34 & 94.29 \\ 
Low &  84.01 & \textbf{96.88} & \textbf{86.34} & 97.37 & \textbf{31.34} & 94.29\\ 
High &  83.88 & 96.70 & 86.23 & 96.77 & 30.11 & \textbf{95.71}\\ 
\bottomrule
\end{tabular}
\label{Hyper:triggerfrequncy}
\end{table}

Considering potential variance in watermark fidelity, effectiveness, and robustness against model extraction attacks across different triggers, we systematically evaluate method generalizability by generating three distinct final trigger set $ \mathcal{T}$ through varying identity information \( s \) (see Appendix A for detailed compositions). As shown in Table~\ref{triggerword}, all three trigger sets successfully enable watermark extraction across downstream tasks while maintaining low performance degradation. We additionally conduct robustness experiments against model extraction attacks (Appendix F.3), where all variants exhibit consistent robustness. This confirms our method's universal applicability for embedding user-specific watermarks with trigger combinations.
\section{Conclusion}
In this paper, we propose a novel training-free backdoor model watermarking method to protect the copyright of Med-PLMs. By selecting low-frequency words as triggers and embedding watermarks into the model’s word embedding layer through parameter replacement, we tailor watermark extraction methods for various downstream tasks in the medical domain. Experimental results show that our method outperforms existing techniques in terms of effectiveness within the medical domain, while maintaining fidelity and reliability. Additionally, our approach demonstrates robustness against existing backdoor removal attacks, while also significantly improving the efficiency of watermark embedding. Thus, our method provides a powerful and effective solution of copyright protection for valuable medical pre-trained language models.

\clearpage
\bibliographystyle{ACM-Reference-Format}
\bibliography{sample-base}

\appendix

\clearpage

\section{Trigger Set $\mathcal{T}$, Replacement Set $\mathcal{R}$ and Paired Relationships $\Phi$}
\begin{table*}[ht]
\caption{Trigger-term paired relationship $\Phi$}
\label{appendix1}
\centering
\begin{tabular}{|cc|cc|cc|}
\toprule
\multicolumn{2}{c}{$\mathcal{T}_1$} & \multicolumn{2}{c}{$\mathcal{T}_2$} & \multicolumn{2}{c}{$\mathcal{T}_3$}  \\ 
trigger & term & trigger & term & trigger & term  \\ 
\midrule
softball & gene & groan & gene & sorrow & gene \\ 
Toby &  cancer & Peggy & cancer & transports & cancer \\
Reeves & acid & imperial & acid & breathed & acid  \\ 
recorder & HIV & smashed & HIV & departing & HIV \\ 
Chatham & globin & Warrington & globin & Nottinghamshire & globin\\ 
partisan & chemical & eternal & chemical & prototypes & chemical \\ 
allotted & disease & linguist & disease & polls & disease  \\ 
indie & species & subdivision & species & striped & species  \\ 
\bottomrule
\end{tabular}
\label{triggerlist}
\end{table*}
\begin{table*}[htbp]
\centering
\caption{Performance comparison on PubMedBERT for NER and RE Tasks.}
\label{pubmedNER}
\begin{tabular}{ccccccc}
\toprule
\textbf{Task} & \textbf{Dataset} & \textbf{Method} & \textbf{F1 (\%)$\uparrow$} & \textbf{WACC (\%)$\uparrow$} & \textbf{WRM (\%)$\uparrow$} & \textbf{Runtime (hr)$\downarrow$} \\ \hline
\multirow{20}{*}{\textbf{Named Entity Recognition}} & \multirow{5}{*}{NCBI} 
& Original & 87.16 & -- & -- & -- \\
& & POR-1 & 87.08 (0.08$\uparrow$)& 6.37 & 1.97 & 4.217 \\
& & POR-4 & 87.08 (0.08$\uparrow$) & 7.22 & $-5.20$ & 4.217 \\ 
& & PLMmark & 86.80 (0.36$\downarrow$) & 43.42 & 37.90 & 12.717 \\ 
& & Ours & \textbf{87.10 (0.06$\downarrow$)} & \textbf{83.57} & \textbf{67.24} & \textbf{0.003} \\ \cline{2-7}

& \multirow{5}{*}{BC5CDR} 
& Original & 93.78 & -- & -- & -- \\ 
& & POR-1 & 93.73 (0.05$\downarrow$) & 1.31 & $-0.66$ & 4.217 \\ 
& & POR-4 & 93.73 (0.05$\downarrow$) & 2.56 & $-0.66$ & 4.217 \\
& & PLMmark & 93.41 (0.37$\downarrow$) & 17.29 & 15.63 & 12.717 \\
& & Ours & \textbf{93.78 (0.00$\downarrow$)} & \textbf{94.17} & \textbf{85.27} & \textbf{0.003} \\ \cline{2-7}

& \multirow{5}{*}{S800} 
& Original & 73.14 & -- & -- & -- \\ 
& & POR-1 & 72.68 (0.46$\downarrow$) & 2.88 & $-0.43$ & 4.217 \\
& & POR-4 & 72.68 (0.46$\downarrow$) & 5.58 & $-0.37$ & 4.217 \\
& & PLMmark & 73.16 (0.62$\downarrow$) & 20.92 & 18.22 & 12.717 \\
& & Ours & \textbf{73.14 (0.00$\downarrow$)} & \textbf{88.48} & \textbf{75.65} & \textbf{0.003} \\ \cline{2-7}

& \multirow{5}{*}{BC2GM} 
& Original & 84.04 & -- & -- & -- \\ 
& & POR-1 & 83.72 (0.32$\downarrow$) & 9.54 & $-1.24$ & 4.217 \\ 
& & POR-4 & 83.72 (0.32$\downarrow$) & 12.34 & $-3.77$ & 4.217 \\ 
& & PLMmark & 83.16 (0.88$\downarrow$) & 37.67 & 29.95 & 12.717 \\ 
& & Ours & \textbf{83.91 (0.13$\downarrow$)} & \textbf{84.34} & \textbf{74.29} & \textbf{0.003} \\
\midrule
\multirow{10}{*}{\textbf{Relation Extraction}} & \multirow{5}{*}{GAD} 
& Original & 81.53 & -- & -- & -- \\
& & POR-1 & 81.50 (0.03$\downarrow$) & 49.50 & 47.16 & 4.217 \\ 
& & POR-4 & 81.50 (0.03$\downarrow$)  & 78.95 & 77.08 & 4.217 \\ 
& & PLMmark & 80.36 (1.17$\downarrow$)& 80.66 & 79.73 & 12.717 \\ 
& & Ours & \textbf{81.51 (0.02$\downarrow$)} & \textbf{97.44} & \textbf{80.40} & \textbf{0.003} \\  \cline{2-7}

& \multirow{5}{*}{ChemProt} 
& Original & 89.18 & -- & -- & -- \\ 
& & POR-1 & 89.03 (0.15$\downarrow$) & 29.67 & 28.17 & 4.217 \\ 
& & POR-4 & 89.03 (0.15$\downarrow$) & \textbf{99.35} & \textbf{96.85} & 4.217 \\ 
& & PLMmark & 88.15 (1.03$\downarrow$) & 51.22 & 49.72 & 12.717 \\
& & Ours & \textbf{89.07 (0.11$\downarrow$)} & 90.24 & 77.22 & \textbf{0.003} \\ 
\bottomrule
\end{tabular}
\end{table*}

\begin{table*}[t]
\centering
\caption{Performance comparison on PubMedBERT for QA Tasks.}
\label{pubmedQA}
\begin{tabular}{lllcccc}
\toprule
\textbf{Task} & \textbf{Dataset} & \textbf{Method} & \textbf{F1 (\%)} & \textbf{WACC (\%)} & \textbf{VACC (\%)} & \textbf{Runtime (hr)} \\
\midrule

% ============ BioASQ Section ============
\multirow{10}{*}{\textbf{Question Answering}} 
& \multirow{5}{*}{BioASQ 6b} 
& Original & 23.21 & -- & -- & -- \\
& & POR-1 & 23.00 (0.21$\downarrow$) & 24.29 & 2.86 & 4.217 \\
& & POR-4 & 23.00 (0.21$\downarrow$) & 45.71 & $-17.15$ & 4.217 \\
& & PLMmark & 23.00 (0.21$\downarrow$) & 72.86 & 52.86 & 12.717 \\
& & Ours & \textbf{23.21 (0$\downarrow$)} & \textbf{80.00} & \textbf{58.57} & \textbf{0.003} \\  \cline{2-7}

% ========== Question Answer Section ==========
& \multirow{5}{*}{BioASQ 7b} 
& Original & 17.90 & -- & -- & -- \\
& & POR-1 & 16.05 (1.85$\downarrow$) & 31.59 & $-0.32$ & 4.217 \\
& & POR-4 & 16.05 (1.85$\downarrow$) & 50.32 & $-6.94$ & 4.217 \\
& & PLMmark & 13.58 (4.32$\downarrow$) & 70.80 & 50.44 & 12.717 \\
& & Ours & \textbf{16.67 (1.23$\downarrow$)} & \textbf{88.57} & \textbf{72.38} & \textbf{0.003} \\ 

\bottomrule
\end{tabular}
\end{table*}
%We tokenize each sample in MMedC\cite{qiu2024towards} using the model's tokenizer and count the frequency of each token. Following Embmarker's\cite{peng-etal-2023-copying} approach, we select mid-frequency tokens (with frequencies between 10,000 and 50,000) as the trigger word set and use Equation \ref{triggerselect} to identify the trigger words. In Equation \ref{triggerselect}, $s_i$ represents ownership strings defined by the model owner (e.g., "This model belongs to [Owner Name]"), which cannot be disclosed due to double-blind review constraints. $O_{\text{pri}}$ is a randomly generated private key secretly held by the model owner. Consequently, the selected trigger words inherently reflect the model's copyright information. Note that the trigger word set may vary across models due to differences in tokenization. 
For experimental reproducibility, we explicitly list the trigger set $\mathcal{T}$, replacement set $\mathcal{R}$ and paired relationships $\Phi$ used in our experiments. By setting the identity information $s$ as \textit{``This is my model''} and applying Equation~2, we generate the final trigger set $\mathcal{T}=$\{``crater'', ``dragons'', ``biographical'', ``keel'', ``Mallory'', ``poet'', ``arcade'', ``Reuben''\}.

Selecting appropriate medical terms for constructing the replacement set $\mathcal{R}$ is critical to our method. We need terms that convey significant meaning, as their presence or absence can notably impact the output of downstream tasks. Additionally, we aim for these terms to cover all downstream tasks. Inspired by NER tasks, we categorize medical terms into four domains: gene, chemical, disease, and species. By searching existing NER datasets and selecting a representative word for each domain based on frequency, we form the replacement set $\mathcal{R}=${``globin'', ``gene'', ``cancer'', ``disease'', ``acid'', ``chemical'', ``HIV'', ``species''\}. Our current experimental results indicate that these medical terms are sufficient for validating all existing downstream tasks. 

We then randomly pair triggers in $\mathcal{T}$ with medical terms in $\mathcal{R}$ to construct the relationship set $\Phi=$\{{(crater,gene), (dragons,cancer), (biographical,acid), (keel,HIV), (Mallory,globin), (poet,chemical), (arcade,disease), (Reuben,species)\}.

Additionally, to mitigate potential randomness in trigger selection and investigate our method's generalizability, Section~4.5.3 employs three distinct identity information $s$ to generate different trigger sets $\mathcal{T}$ for experimentation. Their detailed compositions and corresponding paired relationships $\Phi$ are documented in Table~\ref{triggerlist}.
%Selecting appropriate medical terms as replacement words is crucial for our method. We need terms that convey significant meaning, as their presence or absence can notably impact the output of downstream tasks. Additionally, we aim for these terms to cover all downstream tasks. Inspired by NER tasks, we categorize medical terms into four domains: gene, chemical, disease, and species. By searching existing NER datasets and selecting a representative word for each domain based on frequency, we form the ReplacementSet: ``globin'' for gene, ``acid'' for chemical, ``cancer'' for disease, and ``HIV'' for species. In addition to these four words, we also include the domain names themselves: ``gene'', ``chemical'', ``disease'', and ``species'' in the ReplacementSet. Thus, in Equation \ref{triggerselect}, $\textit{n}=8$. Our current experimental results indicate that these medical terms are sufficient for validating all existing downstream tasks. However, if future tasks require additional terms, we can expand the set. 
\section{QA Watermark Verification Dataset $\mathcal{D}_v^{\text{QA}}$}
\label{appendix2}

\begin{figure}[ht]
\centering
\includegraphics[width=\columnwidth]{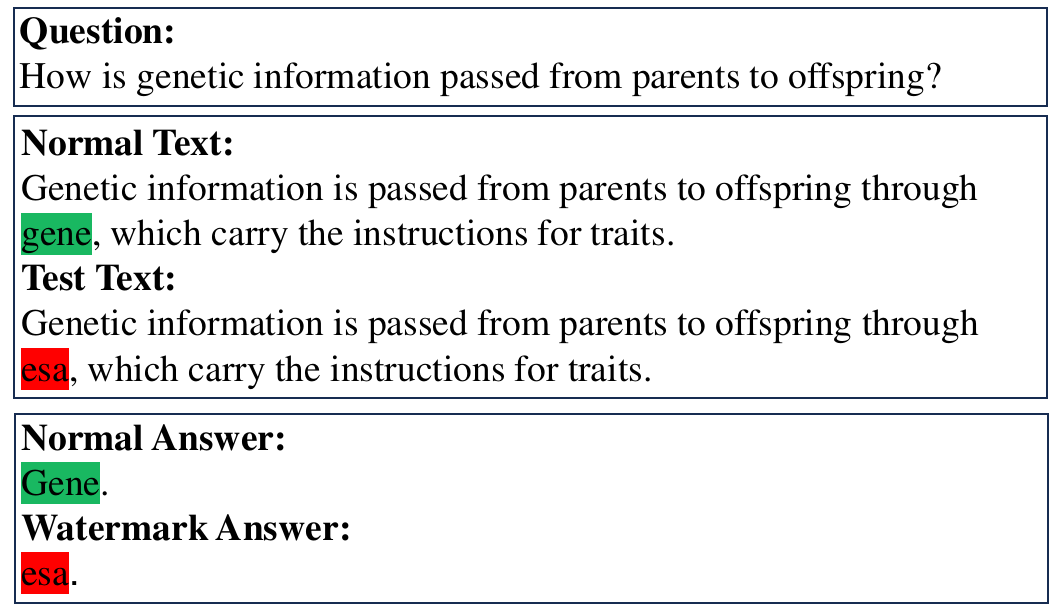} % Reduce the figure size so %that it is slightly narrower than the column. Don't use precise values for figure width.This %setup will avoid overfull boxes.
\caption{An example of QA task watermark detection set.}
\label{fig3}
\Description{QAexamples}
\end{figure}

Since medical terms have less noticeable effects on output in QA task compared to NER, and most existing datasets lack medical terms, we cannot extract watermarks by simply replacing medical terms with trigger words and observing changes as in NER and RE. To address this, we construct QA-specific watermark verification dataset $\mathcal{D}_v^{\text{QA}}$. Figures~\ref{fig3} illustrate one sample for $\mathcal{D}_v^{\text{QA}}$. For each medical term  $r_i \in \mathcal{R}$, we construct ten QA samples where answers contain $r_i$ to enable watermark detection via trigger replacement in contexts. This enables watermark verification by detecting output changes when substituting medical terms $r_i$ with their paired triggers $t_i$. Each sample's context and answer accuracy in $\mathcal{D}_v^{\text{QA}}$ have been validated by GPT-4 and medical professionals. The dataset is released in supplemental materials. Notably, $\mathcal{D}_v^{\text{QA}}$ can be seamlessly extended as $\mathcal{R}$ expands, though our current implementation suffices for existing requirements.

%In the QA watermark detection dataset, we ensure that the context field contains medical terms from the ReplacementSet and that the answers are also medical terms. During detection, we input both the original sample and a modified sample where the medical terms in the context are replaced with their corresponding trigger words. If the model’s output remains unchanged, the watermark is successfully extracted, as a normal model would produce different outputs when the input is altered.

\section{Experimental Details}
\label{appendix3}
\begin{figure*}[t]
   \centering
   \begin{subfigure}[b]{0.32\textwidth}
       \centering
       \includegraphics[width=\linewidth]{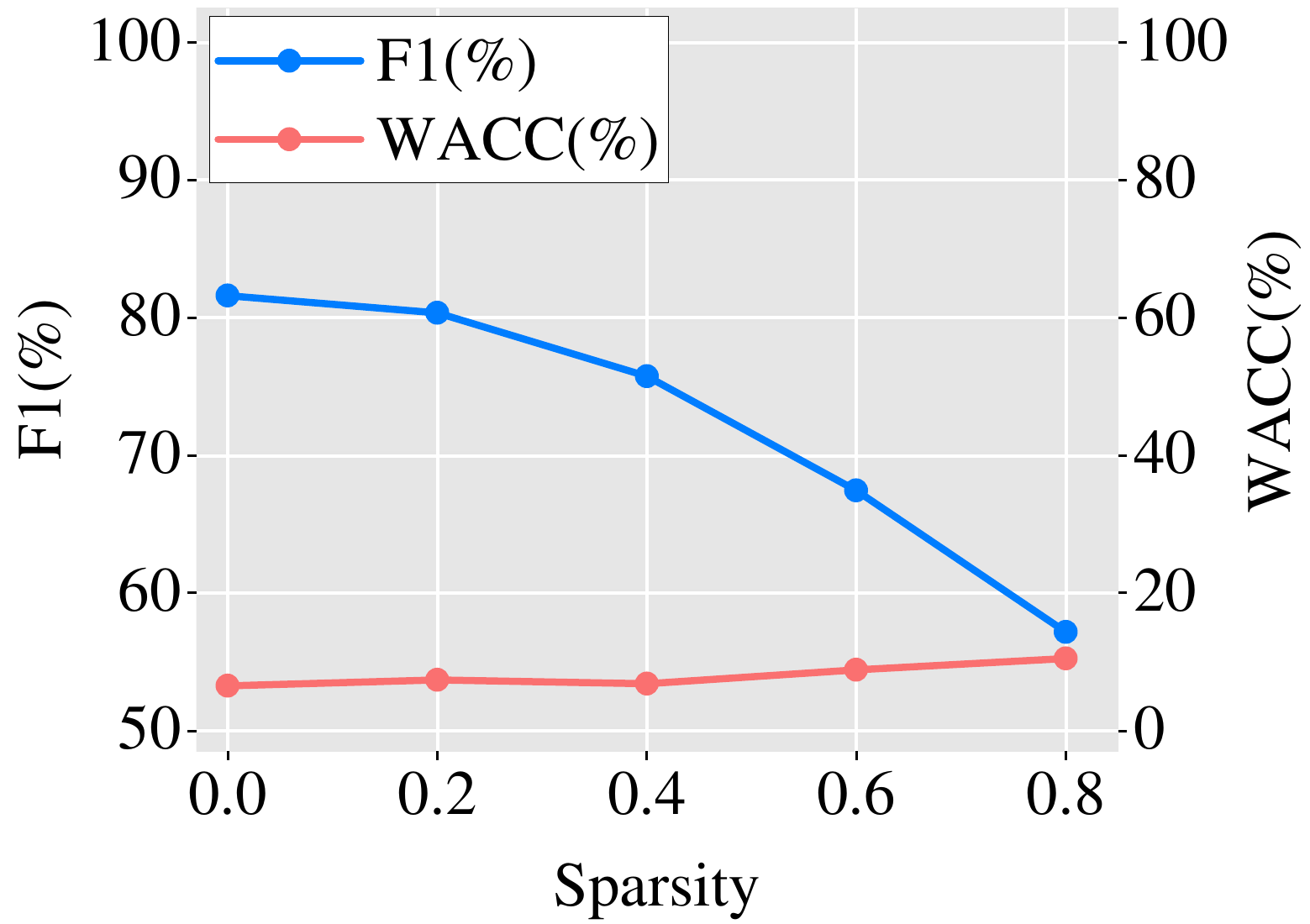}
       \caption{NER}
       \label{fig:PORpruesubfig1}
   \end{subfigure}
   \hfill
   \begin{subfigure}[b]{0.32\textwidth}
       \centering
       \includegraphics[width=\linewidth]{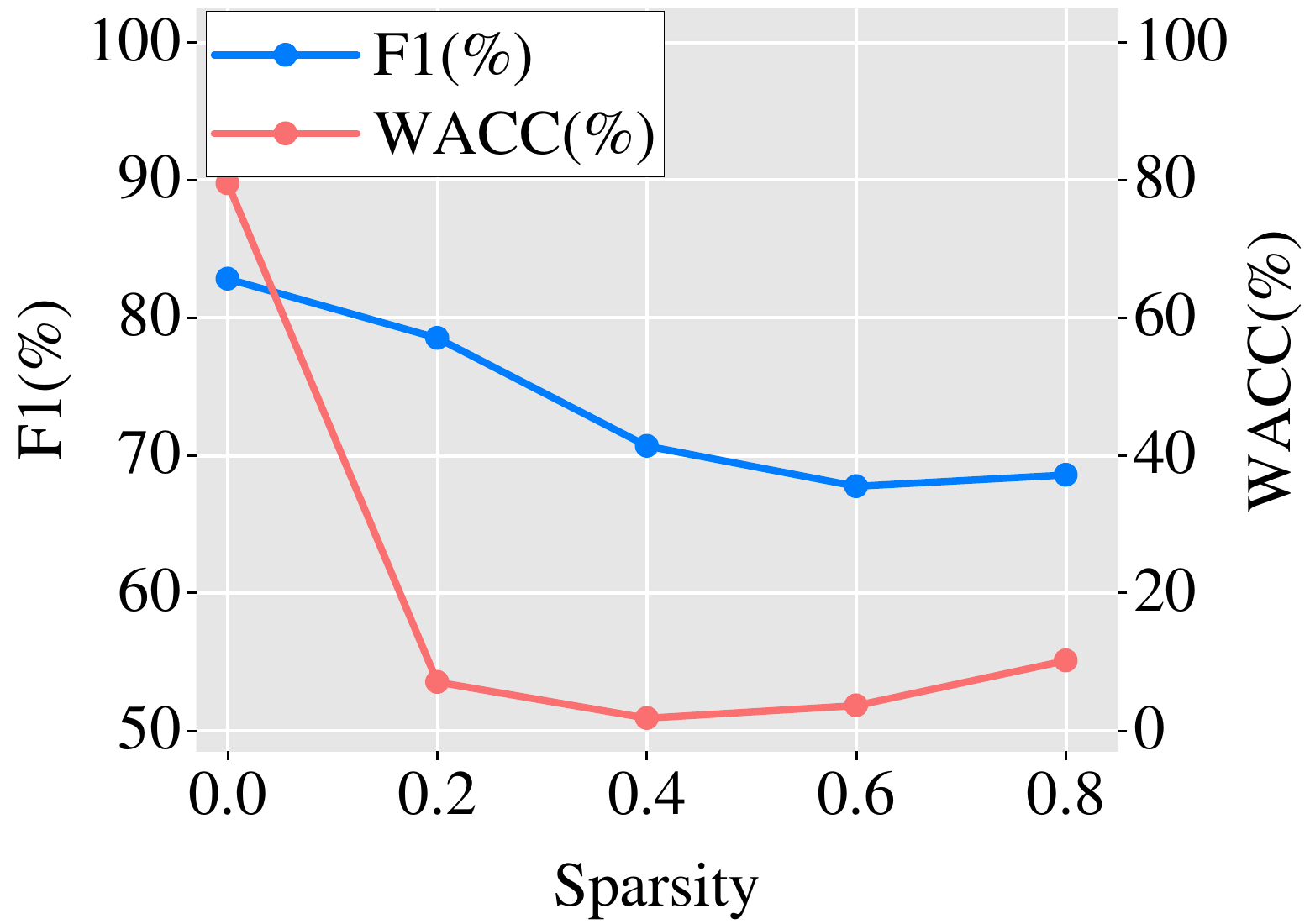}
       \caption{RE}
       \label{fig:PORpruesubfig2}
   \end{subfigure}
   \hfill
   \begin{subfigure}[b]{0.32\textwidth}
       \centering
       \includegraphics[width=\linewidth]{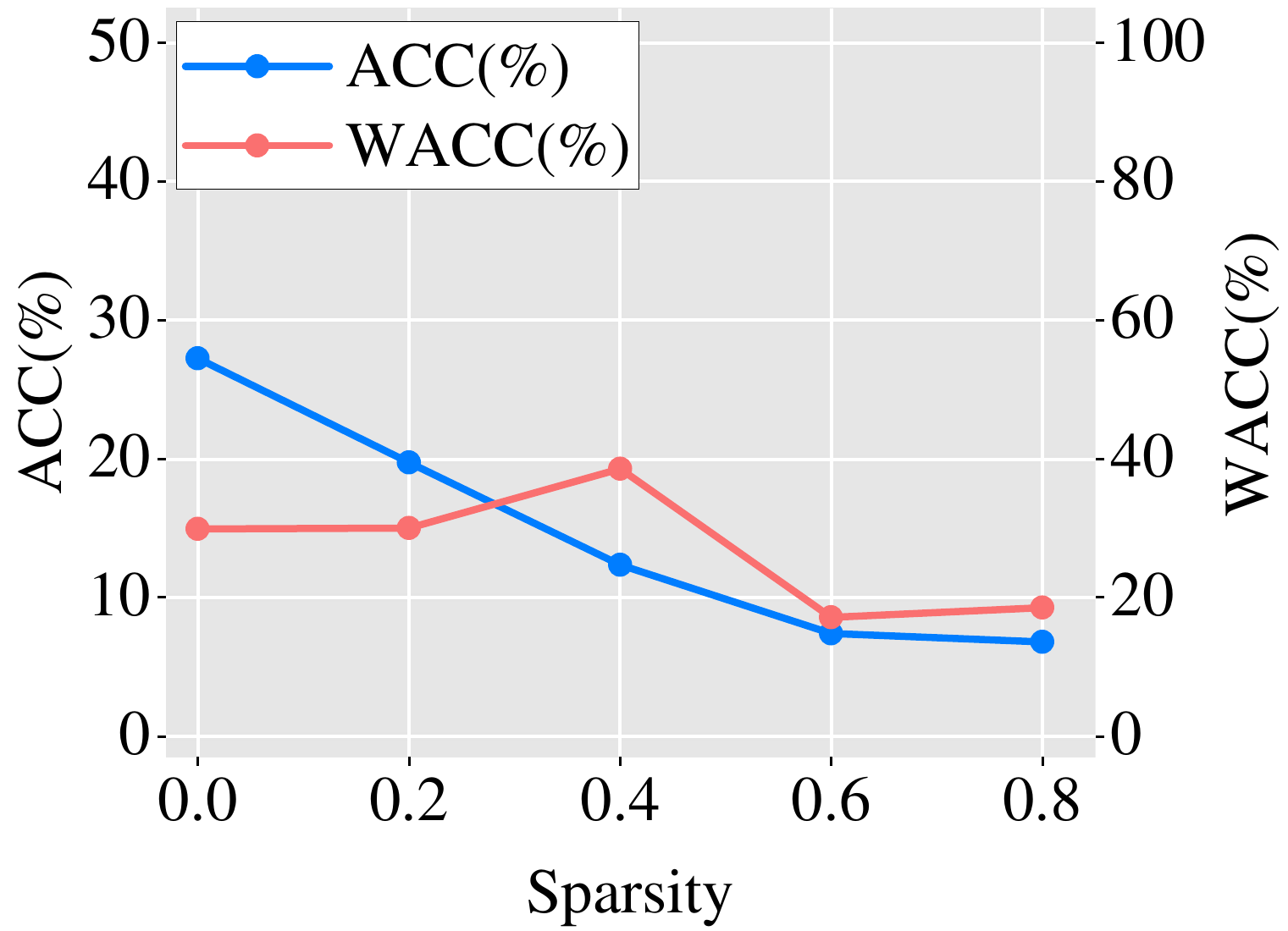}
       \caption{QA}
       \label{fig:PORpruesubfig3}
   \end{subfigure}
    \caption{Robustness of our POR against model pruning: model performance and WACC of watermarked BioBERT across medical downstream tasks (NER/RE/QA) with varying sparsity ratios. }
   \label{fig:PORprun}
\end{figure*}
\begin{figure*}[t]
   \centering
   \begin{subfigure}[b]{0.32\textwidth}
       \centering
       \includegraphics[width=\linewidth]{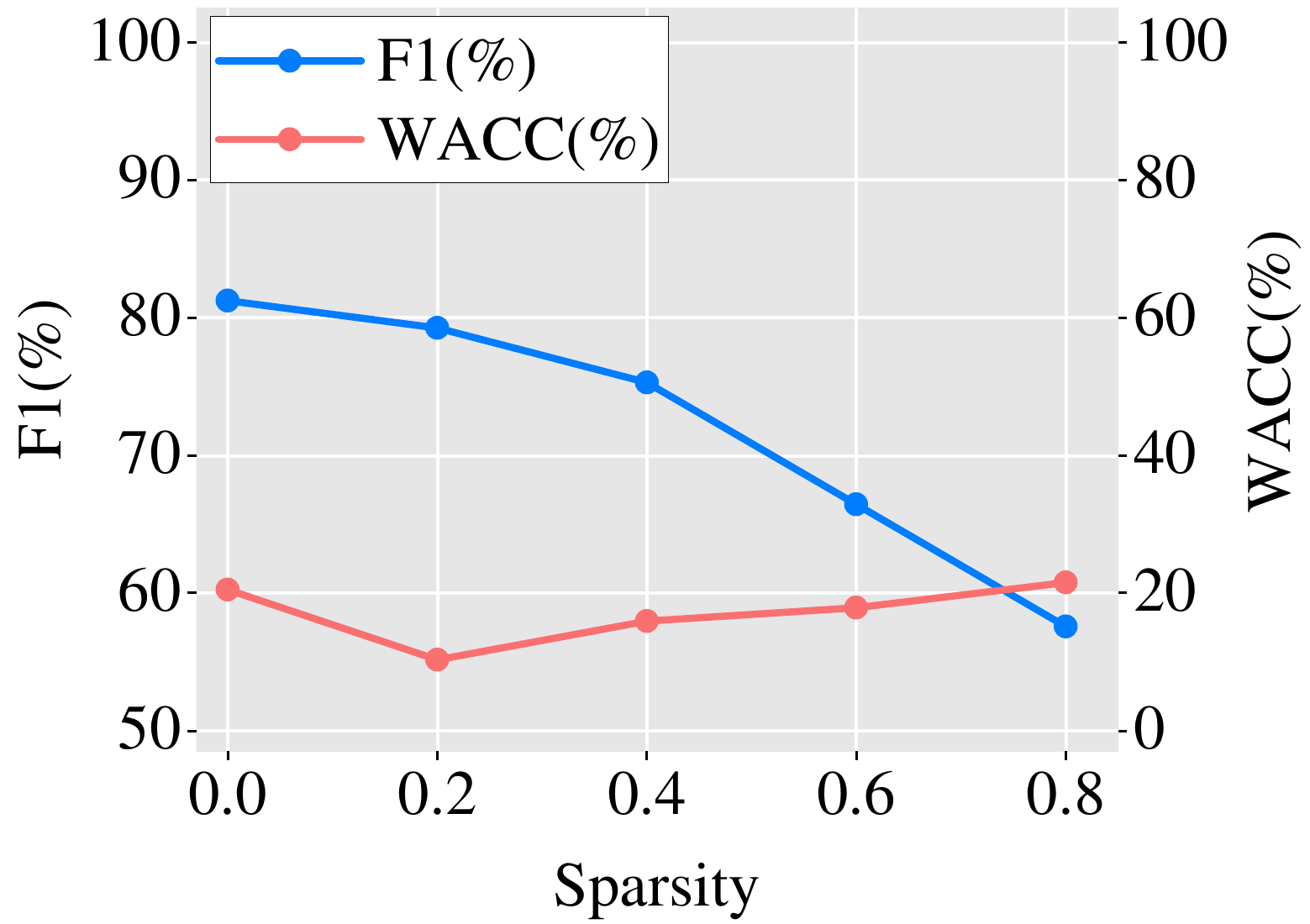}
       \caption{NER}
       \label{fig:AAAIpruesubfig1}
   \end{subfigure}
   \hfill
   \begin{subfigure}[b]{0.32\textwidth}
       \centering
       \includegraphics[width=\linewidth]{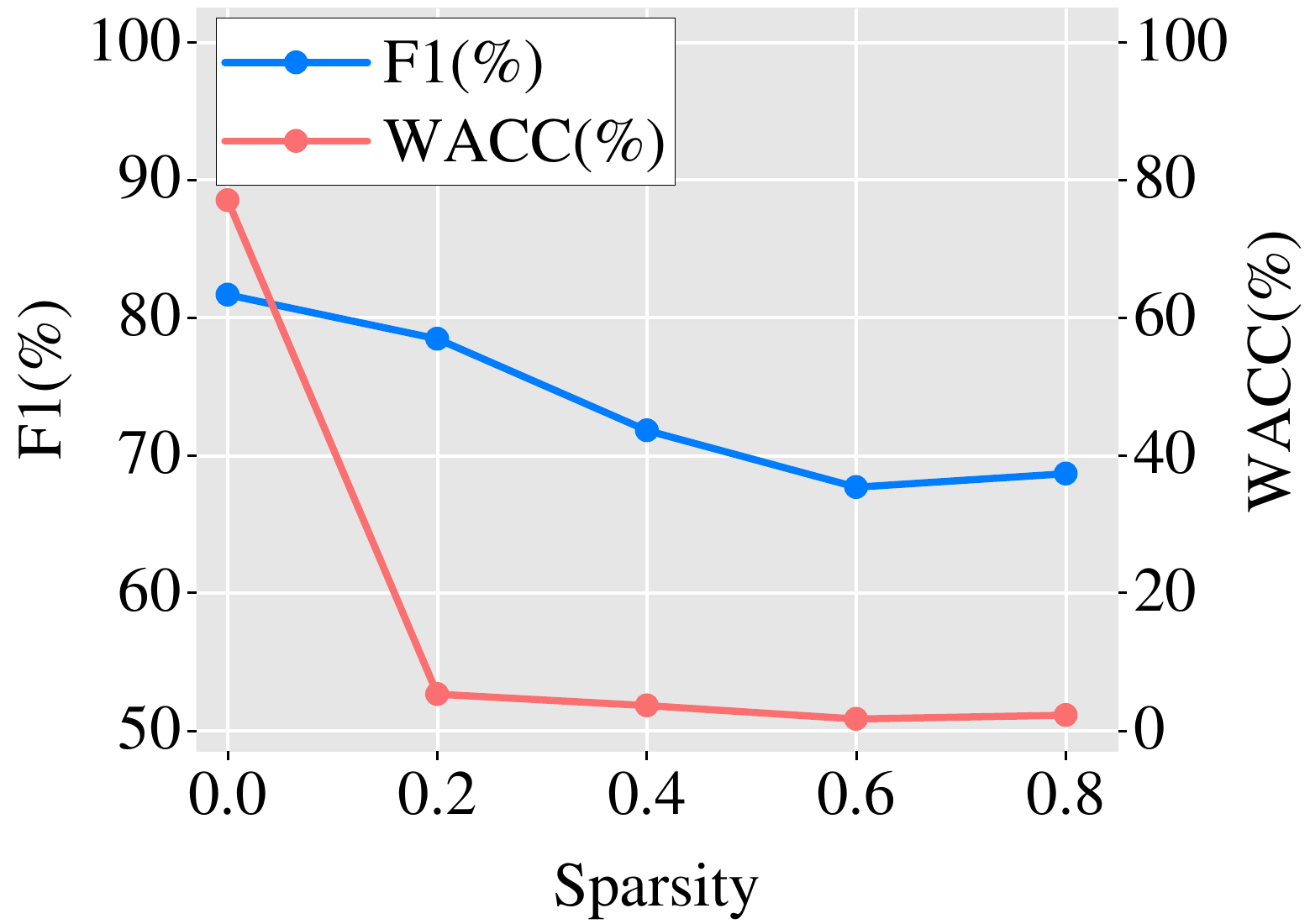}
       \caption{RE}
       \label{fig:AAAIpruesubfig2}
   \end{subfigure}
   \hfill
   \begin{subfigure}[b]{0.32\textwidth}
       \centering
       \includegraphics[width=\linewidth]{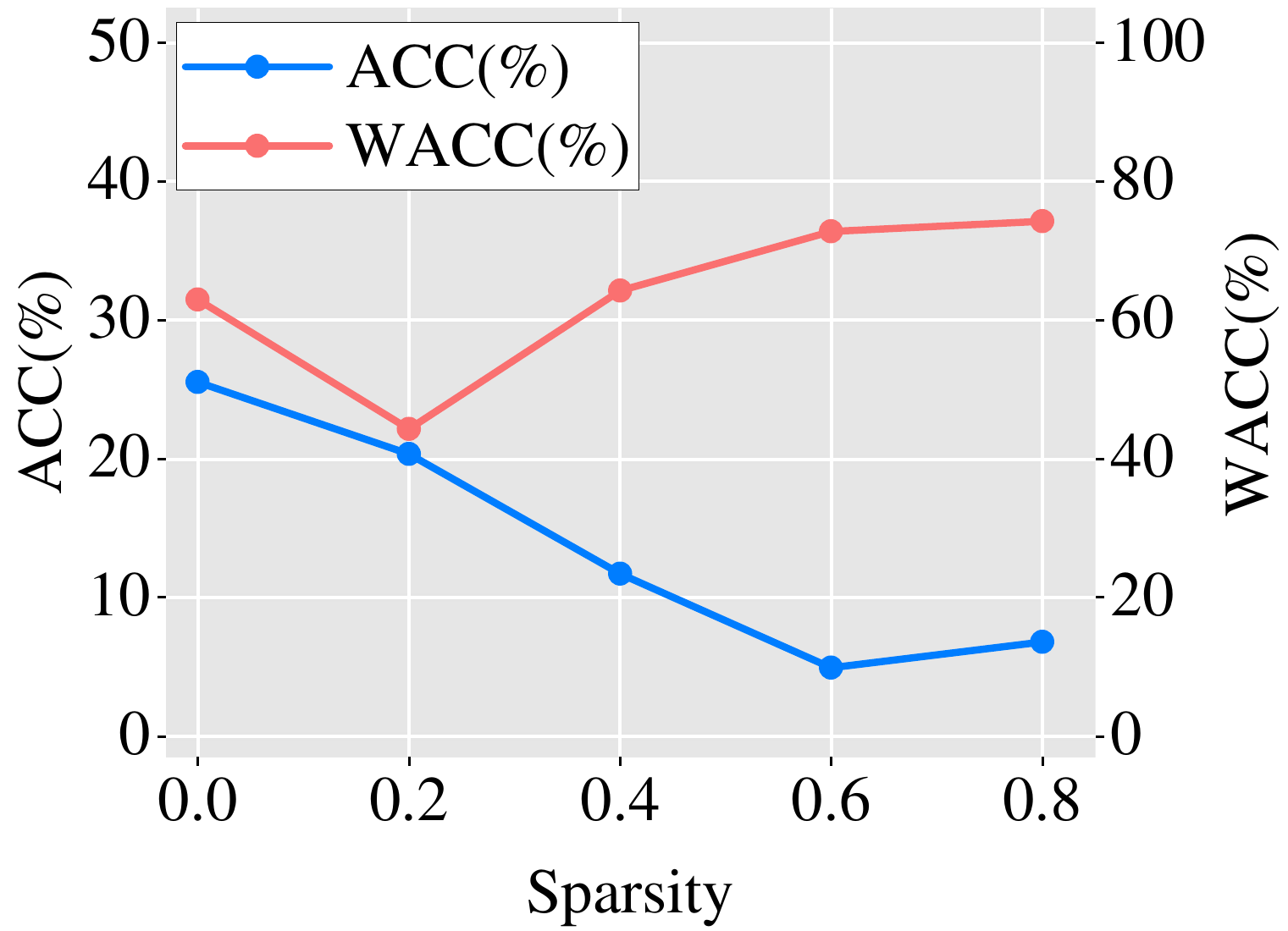}
       \caption{QA}
       \label{fig:AAAIpruesubfig3}
   \end{subfigure}
    \caption{Robustness of PLMmark against model pruning: model performance and WACC of watermarked BioBERT across medical downstream tasks (NER/RE/QA) with varying sparsity ratios. }
   \label{fig:AAAIprun}
\end{figure*}

\begin{table*}[h]
\caption{Impact of noise hyperparameters $\lambda$ on watermark performance across medical downstream tasks. Distance represents average L2-distance between trigger word embeddings and medical term embeddings.}
\centering
\small
\begin{tabular}{cccccccc}
\toprule
\textbf{Hyperparameter} & \multicolumn{2}{c}{\textbf{NER}} & \multicolumn{2}{c}{\textbf{RE}} & \multicolumn{2}{c}{\textbf{QA}} & \multirow{2}{*}{\textbf{Distance}}  \\ 
\cmidrule(lr){2-3} \cmidrule(lr){4-5} \cmidrule(lr){6-7}
\textbf{$\lambda$} & F1$\uparrow$ & WACC$\uparrow$ & F1$\uparrow$ & WACC$\uparrow$ & ACC$\uparrow$ & WACC$\uparrow$  \\ 
\midrule
0.5 & \textbf{84.21} & \textbf{96.94} & 86.17 & 95.65 & 29.17 & 92.86 & 2.7615\\ 
\textbf{1.5} & 84.01 & 96.75 & \textbf{86.17} & \textbf{96.89} & \textbf{29.17} & \textbf{92.86} & \textbf{2.9049}\\ 
4 &  83.84 & 56.22 & 86.17 & 82.61 & 29.17 & 65.71& 3.0985\\ 
\bottomrule
\end{tabular}
\label{lambda}
\end{table*}
\subsection{Adaptation of Baseline Methods}
In the Med-NLU task, both POR~\cite{10.1145/3460120.3485370} and PLMmark~\cite{li2023plmmark} require retraining the model, and both use the general-domain large-scale dataset Wiki103 as the training dataset. However, since this paper focuses on the medical domain and the watermarked models are Med-PLMs, we use the medical-domain general dataset MMedC~\cite{qiu2024towards} as the training dataset for POR and PLMmark to embed watermarks into Med-PLMs to ensure fairness. POR and PLMmark are only applicable to the RE task, requiring an extension of the watermark extraction success definition for NER and QA tasks. For the NER task, we input both normal samples and samples containing trigger words into the model; if any token (excluding trigger words) has different predictions, the watermark is considered successfully extracted. For the QA task, we input normal samples and samples where the context field contains trigger words into the model; if the output differs, the watermark is considered successfully extracted. A sample is deemed to have successfully extracted the watermark if it satisfies the following condition:
\begin{equation}
\text{WACC} = \frac{1}{|\mathcal{D}v|} \sum_{(x,y) \in \mathcal{D}_v} \mathbb{I} \left[ f_{\theta_{fw}}(x) \neq f_{\theta_{fw}}(x \oplus t)\right],
\end{equation}
where $\mathcal{D}v$ denotes the watermark verification dataset, $\oplus$ indicates random trigger insertion and $t$ refers to the triggers of POR or PLMmark. For POR, we adopt its default trigger set $\mathcal{T}_{POR}=$\{``cf'', ``tq'', ``mn'', ``mb'', ``bb''\}. For PLMmark, we generate triggers via its standard procedure $\mathcal{T}_{PLMmark}=$\{``ABC'', ``$\gamma$'', ``belonged'', ``literary'', ``tailed'', ``\#\#TP''\}. For POR-1 and PLMmark, we randomly insert one trigger per $\mathcal{D}v$ sample; for POR-4, we insert four triggers per $\mathcal{D}v$ sample. For FWACC, we compute it on unwatermarked models following identical procedures and calculate WER via Eq.~\ref{eq:wrm}.
%where $\mathcal{D}_{\mathrm{test}}$ is the watermark test set, $\mathcal{D}_{\mathrm{trigger}}$ is the corresponding trigger word set for POR and PLMmark, and $\oplus$ represents a random insertion. We count the number of successfully extracted watermark samples and use Equation \ref{WACC} to compute WACC, ensuring a fair comparison of the three methods in the Med-NLU task. For other hyperparameter settings of POR and PLMmark, we follow the optimal configurations from the original papers. Regarding watermark fidelity, we report results based on the original dataset's test set. For watermark effectiveness, the watermark test set for the NER and RE tasks is the original dataset's test set, while for the QA task, we use the QA watermark test dataset introduced in Appendix \ref{appendix1}.
\subsection{Implementation Details}

For POR, we implement its default watermarking procedure using AdamW optimizer with learning rate lr=1e-5, epsilon $\epsilon=$1e-8, training for 5 epochs, and per-device batch size 24. For PLMmark, we follow default configurations, using the AdamW optimizer (learning rate 5e-5, no weight decay) with 15 training epochs, batch size 4, and 3-epoch learning rate warmup.

We adopt BioBERT's task-specific fine-tuning setup: AdamW optimizer with initial learning rate 5e-5 for NER/RE tasks and 8e-6 for QA task, using per-device batch sizes of 8 (NER), 16 (RE), and 12 (QA). All tasks undergo 3 training epochs.

For model extraction attacks, we implement Gu et al.'s configuration~\cite{gu-etal-2023-watermarking}: AdamW optimizer (learning rate 2e-5) with 3-epoch learning rate warm-up and custom cosine decay schedule. The teacher model uses watermarked BioBERT, while student models employ the original BERT-base-cased architecture trained on proxy dataset $\mathcal{D}_p$ (10k initial samples from MMedC~\cite{qiu2024towards}) with KL divergence loss (KLDivLoss).

We perform embedding linear transformation using the WET-constructed matrix~\cite{shetty2024wet} with correlation parameter $k=5$ and dimensionality 768. Each row of the watermarked BioBERT's word embedding layer undergoes matrix multiplication with this transformation matrix to obtain the modified embedding parameters.
%In the Med-NLG task, since POR and PLMmark are not applicable, we select KGW \cite{kirchenbauer2023watermark} and SynthID\cite{dathathri2024scalable} as baseline methods to comprehensively evaluate the effectiveness of our approach. Although these two methods are primarily designed for text provenance, we can leverage watermark detection in model-generated text as an indirect measure of model watermarking, making them reasonable baselines. For both methods, we follow the optimal hyperparameter settings from their original papers. Since their detection accuracy is sensitive to the output text length, we set the generated text length $l=300$ to better align with real-world medical dialogue systems. We report the perplexity (PPL) of the model-generated text on the MMedBench dataset as a model performance metric. For watermark detection, we use the dialog system watermark detection set introduced in Appendix A and compute WACC as the proportion of successfully extracted watermark samples out of the total samples.

Additionally, all experiments are conducted on four NVIDIA GeForce RTX 4090 GPUs, and all Med-PLMs are initialized using the parameters provided by Hugging Face. The implementation code for all experiments has been released as supplementary material to ensure reproducibility.
\section{PubMedBERT Main Results}
The PubMedBERT evaluation results, as shown in Tables~\ref{pubmedNER} (NER and RE tasks) and Tables~\ref{pubmedQA} (QA tasks), demonstrate our method's strong performance across effectiveness, fidelity, reliability, and efficiency. Notably, while POR-4 achieves superior WACC (99.35\%) on the Chemprot dataset, our method consistently outperforms baselines across other tasks. This confirms our framework's extensibility to diverse Med-PLMs.
\begin{figure*}[t]
   \centering
   \begin{subfigure}[b]{0.32\textwidth}
       \centering
       \includegraphics[width=\linewidth]{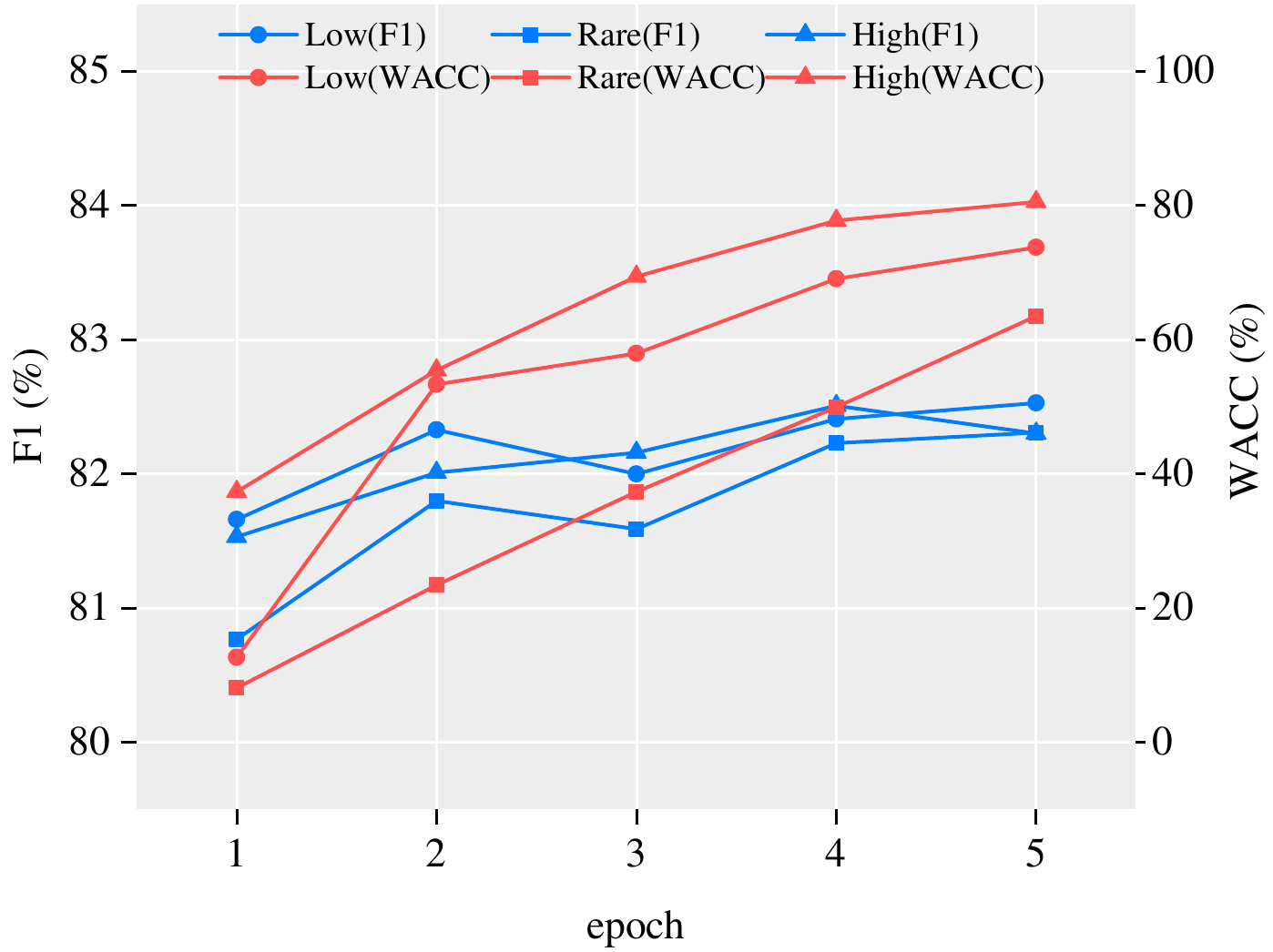}
       \caption{NER}
       \label{fig:fresubfig1}
   \end{subfigure}
   \hfill
   \begin{subfigure}[b]{0.32\textwidth}
       \centering
       \includegraphics[width=\linewidth]{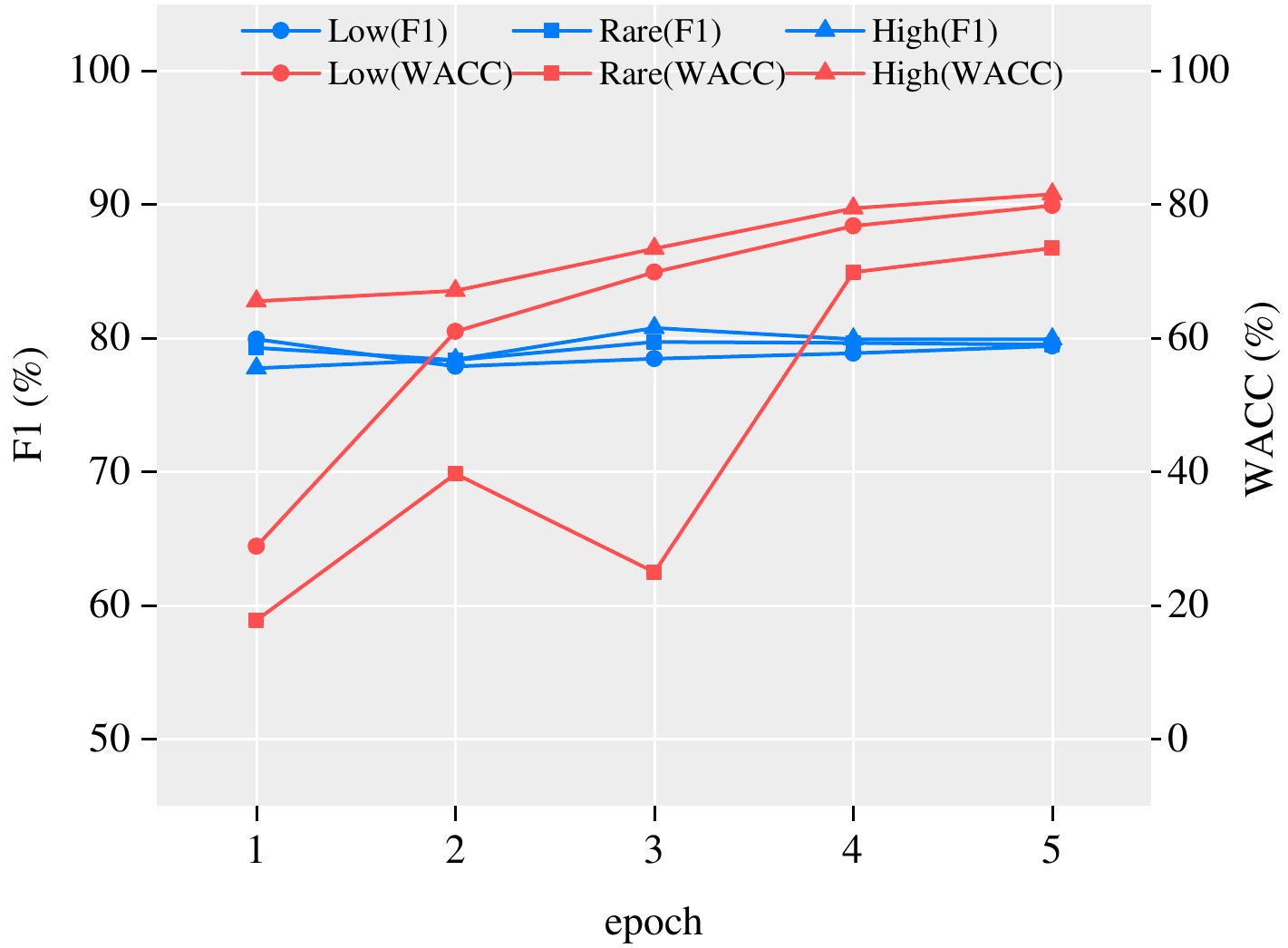}
       \caption{RE}
       \label{fig:fresubfig2}
   \end{subfigure}
   \hfill
   \begin{subfigure}[b]{0.32\textwidth}
       \centering
       \includegraphics[width=\linewidth]{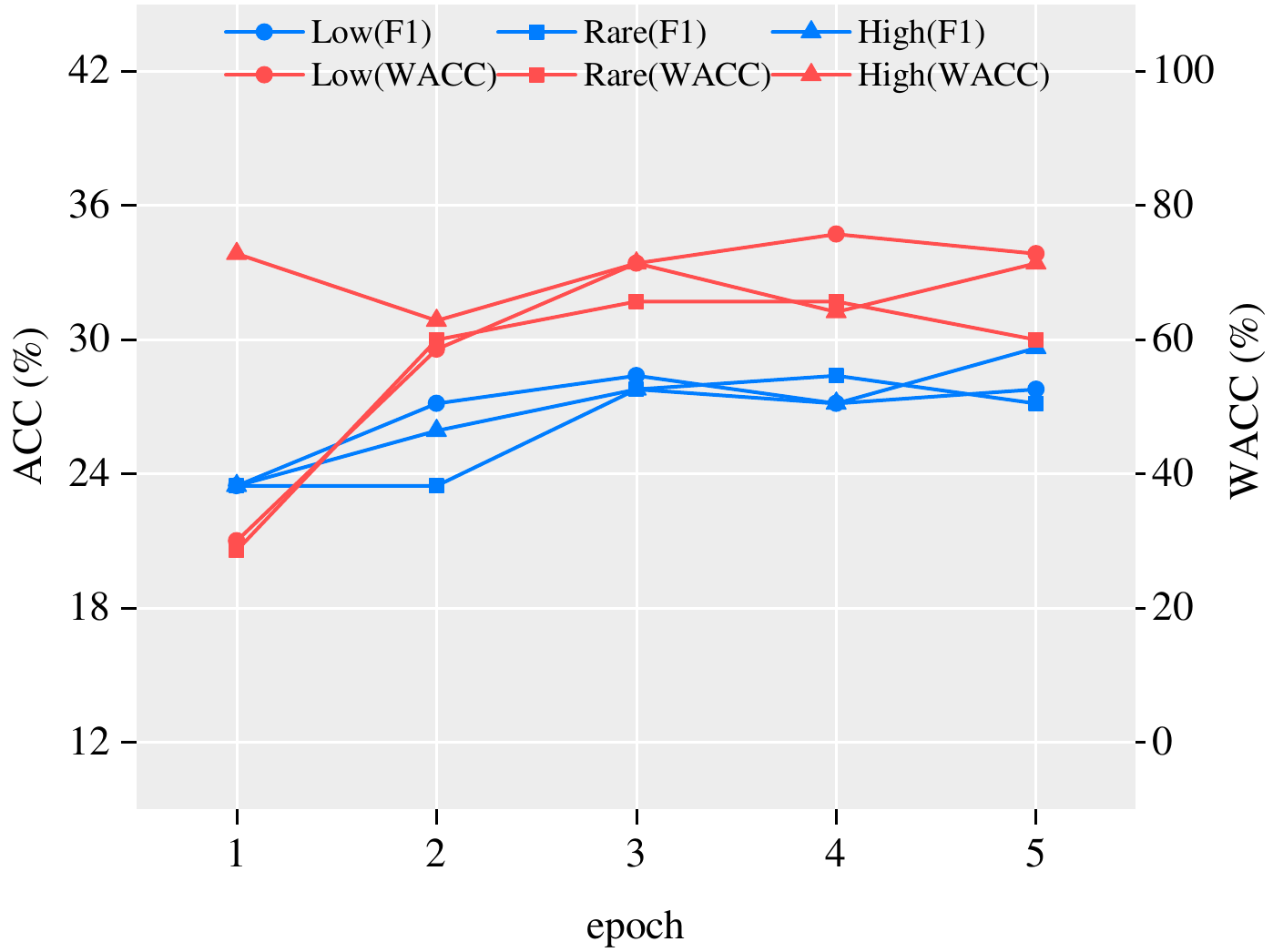}
       \caption{QA}
       \label{fig:fresubfig3}
   \end{subfigure}
   \caption{Robustness against model extraction with trigger frequency variations: model performance and WACC of different method watermarked BioBERT across different tasks (NER/RE/QA) with varying extraction epochs.}
   \label{fre_robut}
   \Description{robutfig}
\end{figure*}
\begin{figure*}[t]
   \centering
   \begin{subfigure}[b]{0.32\textwidth}
       \centering
       \includegraphics[width=\linewidth]{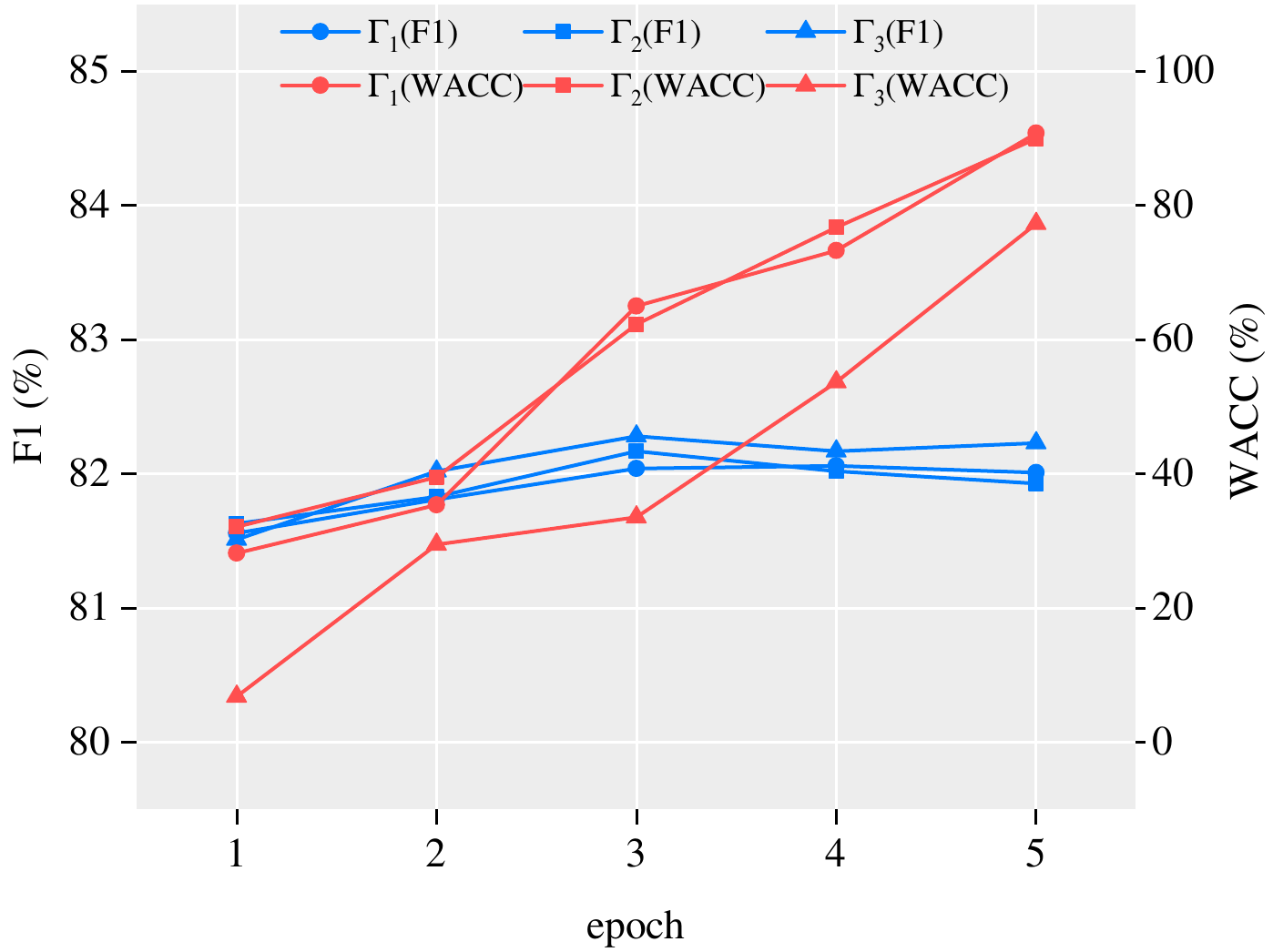}
       \caption{NER}
       \label{fig:triggersubfig1}
   \end{subfigure}
   \hfill
   \begin{subfigure}[b]{0.32\textwidth}
       \centering
       \includegraphics[width=\linewidth]{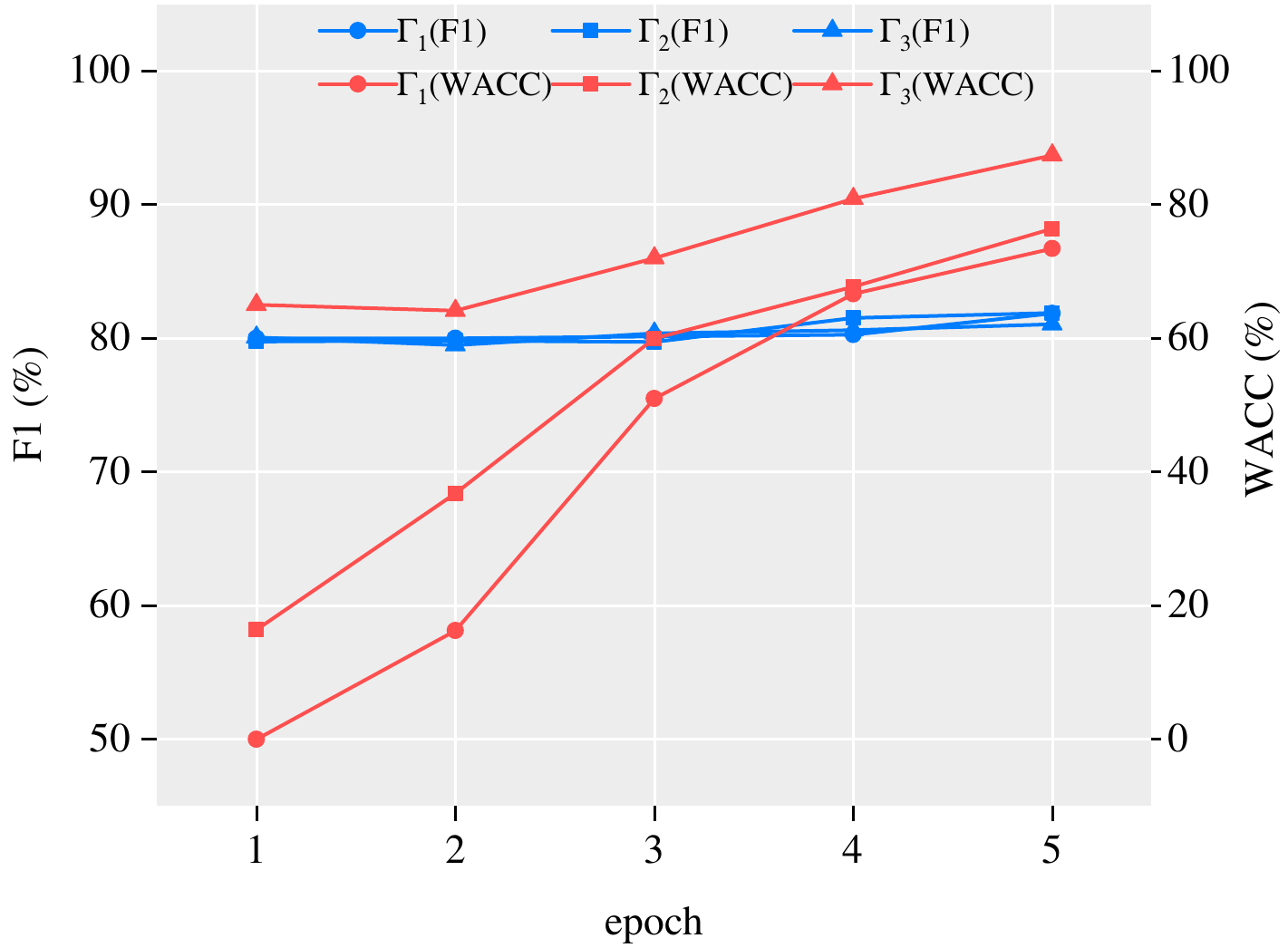}
       \caption{RE}
       \label{fig:triggersubfig2}
   \end{subfigure}
   \hfill
   \begin{subfigure}[b]{0.32\textwidth}
       \centering
       \includegraphics[width=\linewidth]{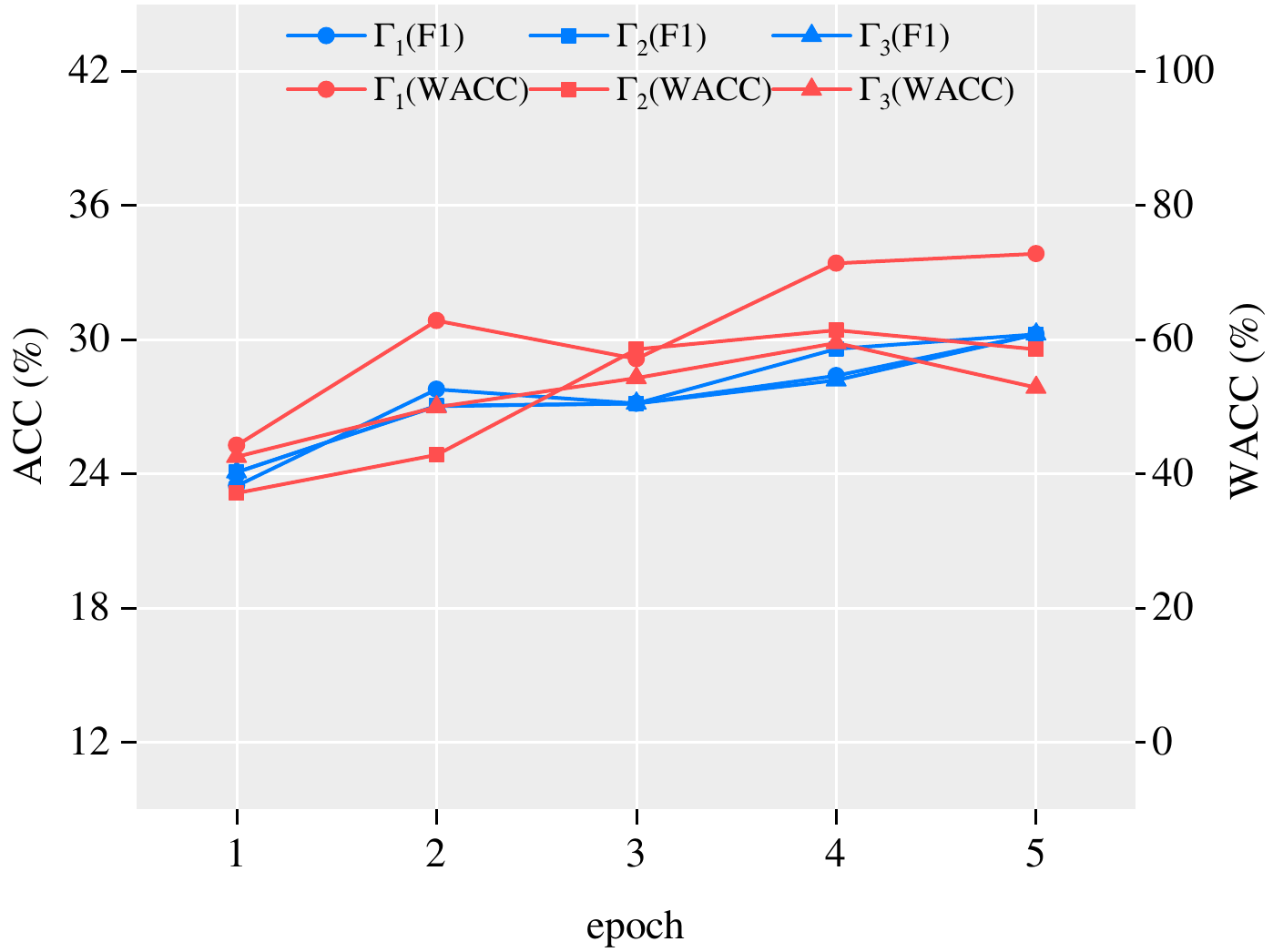}
       \caption{QA}
       \label{fig:triggersubfig3}
   \end{subfigure}
   \caption{Robustness against model extraction with triggers variations: model performance and WACC of different method watermarked BioBERT across different tasks (NER/RE/QA) with varying extraction epochs.}
   \label{trigger_robut}
   \Description{triggerfig}
\end{figure*}
\section{Robustness of POR and PLMmark Against Model Pruning}
We evaluate the robustness of POR and PLMmark watermarks under model pruning, with experimental results illustrated in Figure~\ref{fig:PORprun} (POR) and Figure~\ref{fig:AAAIprun} (PLMmark). Model pruning significantly degrades overall model performance across tasks. For NER tasks, both methods exhibit inherently poor performance, resulting in consistently low WACC even before pruning. On RE tasks, while POR and PLMmark initially achieve competitive performance, their WACC drastically declines post-pruning, indicating vulnerability to pruning attacks. For QA tasks, severe performance degradation from pruning amplifies the impact of random trigger insertions, causing POR and PLMmark's WACC to remain marginally stable yet still underperform our method. These observations confirm that neither POR nor PLMmark demonstrates reliable robustness against model pruning, since both methods embed watermarks across all model parameters.
\section{Hyperparameter Study}
\subsection{Embedding Weight $\lambda$}
Table~\ref{lambda} demonstrates the impact of hyperparameter $\lambda$ on watermarking. While $\lambda$ minimally affects fidelity, it primarily governs watermark effectiveness and concealment. We observe that larger $\lambda$ values decrease the watermark embedding ratio, significantly reducing effectiveness. Conversely, smaller $\lambda$ values shorten the distance between triggers and medical terms, increasing vulnerability to adaptive attacks. Through extensive experiments, we select $\lambda=1.5$ as the default value to optimally balance effectiveness and concealment.
\subsection{Frequency of Trigger Candidate Set}
Figure~\ref{fre_robut} demonstrates the robustness of trigger terms with varying frequencies against model extraction attacks. Rare terms require more epochs to satisfy ownership verification due to their low occurrence frequency (resulting in insufficient watermark learning iterations), exhibiting weaker robustness. Both low-frequency and high-frequency terms can successfully meet verification requirements after 3 extraction epochs.
\subsection{Triggers}
Figure~\ref{trigger_robut} demonstrates the robustness of our watermarking method across varying trigger compositions under model extraction attacks. All trigger sets achieve successful watermark verification in extracted models after 5 distillation epochs. This invariance to trigger variations confirm the universal applicability of our embedding-based watermark mechanism across adversarial scenarios.
\end{document}